\documentclass[letterpaper]{article}
\PassOptionsToPackage{table}{xcolor}
\usepackage{aaai2027}
\usepackage[hyphens]{url}
\usepackage{graphicx}
\usepackage{natbib}
\usepackage{caption}
\usepackage{booktabs}
\usepackage{multirow}
\usepackage{amsmath}
\usepackage{amsfonts}
\usepackage{nicefrac}
\usepackage{subcaption}
\usepackage[most]{tcolorbox}

\newcommand{\methodname}{AHD Agent}

\newcommand{\cref}[1]{%
  \IfBeginWith{#1}{fig:}{Figure~\ref{#1}}{%
    \IfBeginWith{#1}{appendix:}{Appendix~\ref{#1}}{%
      \IfBeginWith{#1}{apx:}{Appendix~\ref{#1}}{\ref{#1}}}}%
}

\title{AHD Agent: Agentic Reinforcement Learning for Automatic Heuristic Design}
\author{
  Haoze Lv\textsuperscript{\rm 1}\equalcontrib,
  Ning Lu\textsuperscript{\rm 1,\rm 2}\equalcontrib,
  Ziang Zhou\textsuperscript{\rm 1},
  Yew-Soon Ong\textsuperscript{\rm 3},
  Shengcai Liu\textsuperscript{\rm 1}\corresponding
}
\affiliations{
  \textsuperscript{\rm 1}Guangdong Provincial Key Laboratory of Brain-Inspired Intelligent Computation,\\
  Department of Computer Science and Engineering, Southern University of Science and Technology,\\
  Shenzhen 518055, China\\
  \textsuperscript{\rm 2}The Hong Kong University of Science and Technology,\\
  \textsuperscript{\rm 3}College of Computing \& Data Science, Nanyang Technological University, Singapore\\
  liusc3@sustech.edu.cn
}

\begin{document}

\maketitle

\begin{abstract}
  Automatic heuristic design (AHD) has emerged as a promising paradigm for solving NP-hard combinatorial optimization problems (COPs).
  Recent works show that large language models (LLMs), when integrated into well-designed frameworks (i.e., LLM-AHD), can autonomously discover high-performing heuristics.
  However, existing LLM-AHD frameworks typically treat LLMs as passive generators within fixed workflows, where the model generates heuristics from manually designed, limited context.
  Such context may fail to capture state-dependent information (e.g., specific failure modes), leading to inefficient trial-and-error exploration.
  To overcome these limitations, we propose \textbf{\methodname{}}, a novel tool-integrated, multi-turn framework that empowers LLMs to proactively decide whether to generate heuristics or invoke tools to retrieve targeted evidence from the solving environment.
  To effectively train such a dynamic decision-making agent, we introduce an agentic reinforcement learning (RL) system, which leverages a novel environment synthesis pipeline to optimize a compact model's generalizable AHD capabilities.
  Experiments across eight diverse domains, including four held-out tasks, demonstrate that our 4B-parameter agent matches or surpasses state-of-the-art baselines using much larger models, while requiring significantly fewer evaluations.
  Model and inference scaling analysis further reveals that \methodname{} offers an effective trajectory toward truly autonomous heuristic design.
\end{abstract}

\begin{links}
  \link{Code}{https://github.com/Antoniano1963/AHD-Agent}
\end{links}

\section{Introduction}

NP-hard combinatorial optimization problems (COPs) are fundamental to many real-world systems, including transportation, planning, and decision-making~\cite{TSPAP:matai2010traveling,FlowshopAP:rajendran1993heuristic}.
Solving these problems efficiently relies heavily on well-designed heuristics~\cite{Intro1COPAppliaction:desale2015heuristic}.
Traditionally, heuristic design is a highly manual and time-consuming process that requires expert analysis of solver behavior and extensive trial and error.
To mitigate these limitations, automatic heuristic design (AHD) has emerged as a promising paradigm for heuristic generation~\cite{HHSurvey2010:burke2010classification}. 
However, traditional AHD approaches, such as genetic programming (GP), still rely heavily on expert-designed components~\cite{HH8:langdon2013foundations,GPAP1:mei2022explainable}.

\begin{figure}[t]
\centering
\includegraphics[width=\linewidth]{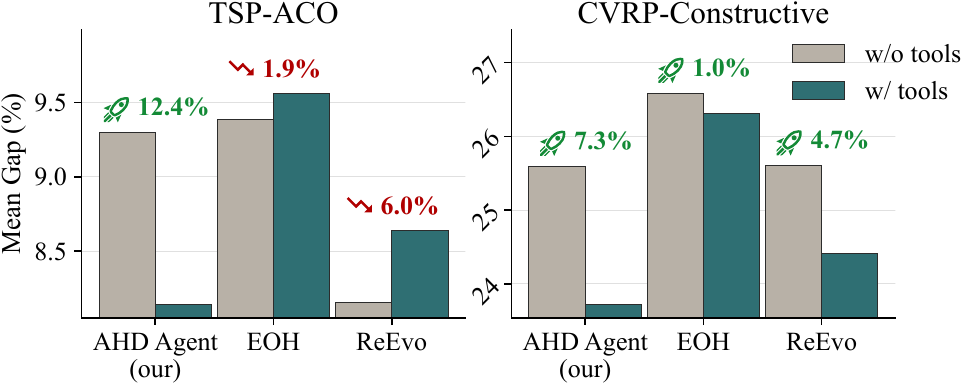}
\caption{\textbf{Tools help \methodname{} more than fixed-workflow LLM-based AHD.} Mean validation gaps are reported under DeepSeek-V4-Flash~\cite{Deepseek-V4:deepseekai2026deepseekv4}. 
For EoH and ReEvo, all tools are called at each LLM-generation step. 
Details are shown in Appendix D.7.
}
\label{fig:deepseek-v4-tool-ablation-mean-gap}
\end{figure}

Recently, large language models (LLMs) have been introduced into AHD as heuristic generators within evolutionary computing (EC) frameworks~\cite{FunSearch:romeraparedes2024mathematical,AlphaEvolve:journals/corr/abs-2506-13131}. 
In these frameworks, LLMs generate new heuristics based on candidates selected according to predefined rules.
The generated heuristics are then evaluated, forming a generation–evaluation feedback loop.

\begin{figure*}[t]
\centering
\includegraphics[width=0.9\textwidth]{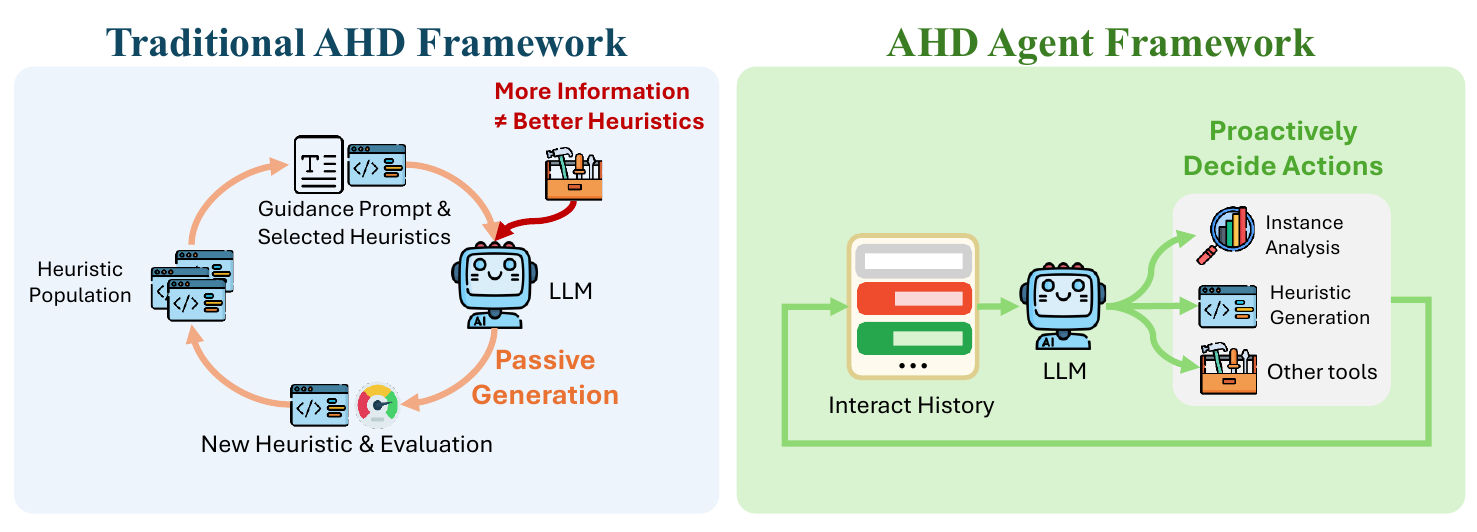}
\caption{\textbf{Traditional LLM-based AHD vs. our \methodname{}.} Traditional AHD places the LLM inside a fixed loop.  \methodname{} enables the LLM to design heuristics by actively calling tools, generating candidates, and performing evaluations. 
}
\label{fig:method-compare}
\end{figure*}

However, existing LLM-based AHD frameworks (e.g., EOH~\cite{EOH:liu2024evolution}, ReEvo~\cite{ReEvo:ye2024reevo}) still face key limitations.
As shown in \cref{fig:method-compare}, they treat LLMs as passive heuristic generators within fixed workflows. These workflows rely on manually designed and limited context (e.g., crossover based on top heuristic~\cite{EOH:liu2024evolution}), which may fail to capture information needed at a specific design step, such as the failure modes of prior heuristics.
As a result, the model cannot identify information gaps or retrieve targeted evidence, and instead relies on inefficient trial-and-error generation.
Our preliminary study in \cref{fig:deepseek-v4-tool-ablation-mean-gap} further shows that simply providing all available information (tools) to LLMs within these fixed workflows brings limited gains and may even hurt performance, suggesting that the key challenge is not information availability alone, but the lack of state-dependent mechanisms for acquiring and using relevant information.
Additionally, existing frameworks typically use general-purpose LLMs that are not specifically aligned for AHD, leading to costly trial-and-error search.

To overcome these limitations, we propose \textbf{\methodname{}}, the first tool-integrated multi-turn framework for LLM-based AHD.
Instead of following a fixed pipeline, \methodname{} enables LLMs to proactively decide whether to generate heuristics or use tools to retrieve relevant information.
This enables the model to adapt its design strategy based on intermediate feedback, like evaluation results and tool outputs.
Building on \methodname{}, we further develop an agentic reinforcement learning (RL) system that optimizes the base model with GRPO~\cite{wang2025ragen} to improve its generalizable AHD capability.
We introduce an AHD RL environment synthesis pipeline that constructs diverse training environments by varying evaluation instances, solver backbones, and initial heuristics.
The RL-trained agent matches or surpasses traditional LLM-based AHD frameworks with substantially fewer heuristic evaluations.
Model and inference scaling further reveal the potential of our framework, suggesting an effective and efficient path toward LLM-based AHD.

Our contributions can be summarized as follows:

\begin{itemize}
    \item We introduce \textbf{\methodname{}}, the first tool-integrated multi-turn framework for LLM-based AHD that enables proactive, state-dependent tool use for heuristic generation, rather than following fixed workflows with static context.
    
    \item We develop an agentic RL system with AHD environment synthesis pipeline and multi-domain joint training, substantially improving the model's generalizable AHD capability across diverse settings.
    
    \item We conduct extensive experiments across eight evaluation settings spanning diverse problem domains, instance scales, and solver backbones. Our 4B-parameter agent outperforms baselines using larger models and exhibits strong generalization, establishing \methodname{} as a competitive and efficient alternative to conventional methods.

\end{itemize}

\section{Related Work}
\textbf{LLM-based AHD.} Recent LLM-based AHD methods use LLMs to generate, evaluate, and refine heuristic code in feedback-driven search loops. FunSearch~\cite{FunSearch:romeraparedes2024mathematical} and EOH~\cite{EOH:liu2024evolution} established this paradigm, followed by fixed-workflow extensions based on reflection or tree search~\cite{ReEvo:ye2024reevo,MCTS-AHD:zheng2025monte,EOH-S:liu2026eoh,HSEVO:dat2025hsevo,MOH:shi2026generalizable,PathWise:gungordu2026pathwise, GLNS:zhao2026g, TIDE:chen2026tide}. Other studies apply LLM-AHD to routing, scheduling, MILP, SAT, and related problems~\citep{VRPAgent:hottung2025vrpagent,LLMFS:li2025llm,DSEvo:zhang2025dhevo,SAT:chen2025dasathco,AgenticVRP:zhang2026afl}. However, most methods still prescribe the search procedure externally, leaving the LLM mainly as a candidate generator.

\textbf{RL for AHD.} Recent studies use RL to improve LLM-based heuristic generation, either within fixed workflows such as CALM or for specific solvers and problem families~\citep{EvoTune:surina2025algorithm,RFTHGS:zhu2026refining,CALM:huang2025calm}. In contrast, \methodname{} learns a transferable heuristic-design policy from multi-domain RL training. The learned policy controls the multi-turn design process itself: it decides when to evaluate, which tools to invoke, and how to revise candidate heuristics based on feedback. Our experiments show that this policy generalizes to unseen problem families and transfers across evaluation protocols.

\textbf{LLM agents and reinforcement learning.} LLM agents have been applied to program generation, device operation, gameplay, and robotic control~\cite{zhang2024codeagent,zhang2023you,wang2023voyager,brohan2023rt}. Early systems mainly used frozen models with prompting strategies such as ReAct and Reflexion, often augmented with memory, retrieval, and tools~\cite{yao2023react,shinn2024reflexion,schick2023toolformer}. Recent work increasingly uses supervised fine-tuning or RL to learn interaction policies directly from environmental feedback~\cite{lu2026policy, wu2026train, gou2026reasoningaligned, wang2026EC}. Representative RL methods span value-based DQN~\cite{mnih2015human}, policy optimization with PPO~\cite{schulman2017proximal}, and recent LLM-agent designs based on MCTS~\cite{silver2017mastering}, RLOO~\cite{kool2019buy}, and trajectory-level GRPO~\cite{wang2025ragen}.

\section{Methodology}
\label{sec:methodology}

We propose \textbf{\methodname{}}, a tool-integrated, multi-turn framework for LLM-based AHD. Unlike fixed-workflow LLM-AHD methods, \methodname{} treats the LLM as the decision-making agent in a multi-turn design process. We first formulate the problem and describe the \methodname{} interaction protocol and tool set. We then present the agentic RL training process, including the AHD environment synthesis pipeline and cross-domain training.


\subsection{The \textit{\methodname{}} Framework}
\label{subsec:agentic-heuristic-designer}

\subsubsection{Formulation}
\label{subsubsec:ahd-formulation}

Let \(\mathcal{I}\) denote the instance space of a target optimization problem, and let \(\mathcal{D}_{\mathrm{design}}\subset\mathcal{I}\) be the design set visible during one heuristic-design episode. The goal of AHD is to construct a heuristic \(h\), represented in executable code form. The evaluator runs \(h\) under the specified solver setting on \(\mathcal{D}_{\mathrm{design}}\) and returns a scalar score \(\operatorname{Score}(h;\mathcal{D}_{\mathrm{design}})\), together with execution feedback. We normalize \(\operatorname{Score}\) so that larger values are always better.

In the \methodname{} framework, the AHD process is formulated as a finite-horizon Markov Decision Process (MDP) \(\mathcal{M}=(\mathcal{S},\mathcal{A},P,R)\). Here \(\mathcal{S}\) represents the state space, where each state \(s_t\) can be an observation sequence or interaction history; \(\mathcal{A}\) is the token-level action space, covering heuristic generation/evaluation, tool calls, and final responses; \(P\) and \(R\) denote the transition dynamics and reward generation process, respectively. The initial state \(s_0\) contains the problem description and a seed heuristic \(h_0\). At each time step \(t\), the agent policy \(\pi_\theta\) generates an action conditioned on the current state \(s_t\) and the interaction history \(\tau_{<t}\), after which the environment returns a reward and a new state:
\[
a_t\sim\pi_\theta(\cdot\mid s_t,\tau_{<t}),
r_t=R(s_t,a_t),
s_{t+1}\sim P(\cdot\mid s_t,a_t).
\]
Here \(\tau_{<t}=\{s_0,a_0,r_0,\ldots,s_{t-1},a_{t-1},r_{t-1}\}\) denotes the interaction history. This interactive process continues for a maximum horizon \(K\) or until a final response is produced. We denote the final response as \(h_{\mathrm{final}}\), where the model returns a heuristic in a predefined format as the final answer.
We set intermediate rewards to zero, and the episode return is determined only by the final heuristic score, i.e., \(R(\tau)=\operatorname{Score}(h_{\mathrm{final}};\mathcal{D}_{\mathrm{design}})\).

\begin{figure*}[t]
\centering
\includegraphics[width=0.9\textwidth]{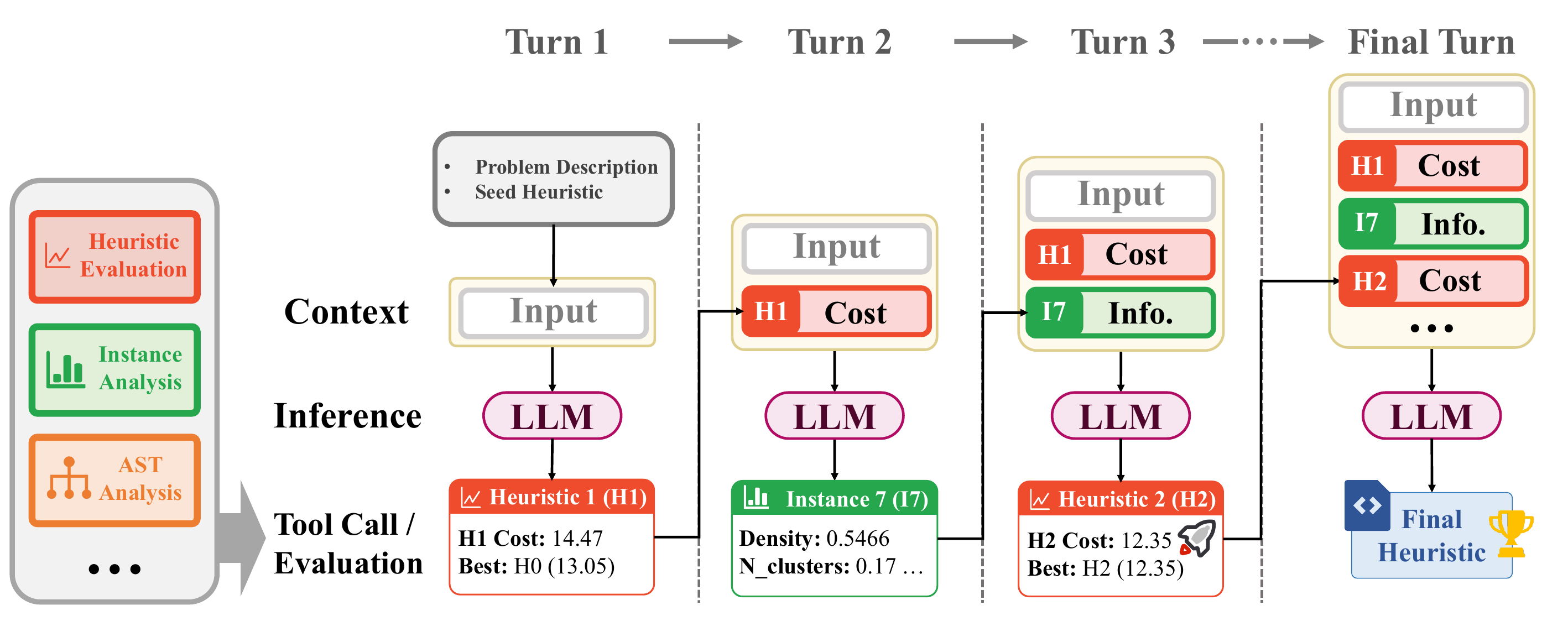}
\caption{\textbf{Demonstration of \methodname{} workflow.} Given a problem description, a seed heuristic, and a set of tools, the model iteratively decides its next action based on the session history, which records all previous interactions. At each turn, it can call tools, generate and evaluate heuristics, and finally return the final heuristic.}
\label{fig:multi_turn_tool_loop}
\end{figure*}

\subsubsection{Agent Loop}

Figure~\ref{fig:multi_turn_tool_loop} illustrates how \methodname{} conducts a heuristic-design episode through multi-turn decision making.
The episode starts from the initial state \(s_0\), which contains the problem description and a seed heuristic \(h_0\).
At the first turn, the agent generates a candidate heuristic and submits it to the evaluator. The resulting execution feedback and objective score are then recorded, updating both the state $s$ and the interaction history $\tau$. In a later turn, the agent calls a tool to inspect specific properties of \(\mathcal{D}_{\mathrm{design}}\). After multiple turns, the agent produces a final answer, and the framework returns the final heuristic \(h_{\mathrm{final}}\).

In practical evaluation, the design loop is constrained by a maximum evaluator-call budget \(B\).
Evaluator submissions consume this budget, while diagnostic tool calls do not.
When the remaining evaluator budget approaches the limit, the framework reminds the agent to stop further exploration and output a final heuristic.

\subsubsection{Tool Library}
\label{subsubsec:tool-library}

The tool library provides a set of callable tools through which the LLM Agent can acquire information during heuristic design.
At each turn of a design run, \methodname{} decides whether to invoke a tool according to the current design state.
If a tool call is needed, the agent further selects which tool to use, enabling it to acquire information adaptively during the search process.
Moreover, this modular design makes the framework flexible: adding or removing a tool only changes the information that the agent can actively request, without requiring a manually redesigned search workflow. All tools in the tool library are design-set-only: they can inspect the design dataset \(\mathcal{D}_{\mathrm{design}}\), the current interaction history, and the evaluator feedback already produced during the design run, but they never reveal the validation dataset. The current tool library contains two read-only diagnostic tool groups.

\textbf{Instance-analysis tools.}
Instance-analysis tools provide three query modes to expose structural information about \(\mathcal{D}_{\mathrm{design}}\): \texttt{summary}, \texttt{single\_instance}, and \texttt{contrastive\_pair}.
The \texttt{summary} mode returns dataset-level statistics, including scale, spatial, nearest-neighbor, density, and domain-specific features when available.
The \texttt{single\_instance} mode inspects one selected instance, while the \texttt{contrastive\_pair} mode contrasts the most dissimilar pair of design instances to reveal structural heterogeneity.
By selecting the appropriate mode, the agent can acquire instance-level information when needed, e.g., to decide whether a heuristic should account for clustered nodes, capacity-demand patterns, or heterogeneous instance scales.

\textbf{AST tools.}
AST tools compare a candidate program against previous attempts in the same session at the abstract-syntax-tree level and quantify its structural novelty. If a candidate is structurally close to previously evaluated code, the agent can revise it before committing an evaluator call. Moreover, when recent attempts exhibit low structural novelty, the agent can use AST-derived information to choose larger structural revisions, reducing the risk of being trapped in local refinement loops during heuristic search.

Overall, the tool library is designed to support active information acquisition rather than to inject hand-designed heuristic rules. The tools expose information about the design dataset and code novelty, while the LLM Agent remains responsible for deciding when to call them and how to translate their observations into heuristic revisions.  Detailed tool interfaces and outputs are provided in Appendix B.1.

\subsection{Agentic RL Training}
\label{subsec:agentic-rl}
An AHD environment consists of an exploration setup and an evaluation protocol.
The exploration setup is specified by the seed heuristic, when provided, which defines the initial state of the agent's search.
The evaluation protocol is specified by the target problem, design set, and solver backbone, which determine the feedback for each generated heuristic.
This definition makes environment synthesis automatic: instances are generated, and candidate heuristics are evaluated by execution.
We synthesize environments by varying three elements that shape the feedback: seed heuristics, design-set instances, and evaluation solver backbones.

\subsubsection{AHD Environment Synthesis Pipeline}

RL training requires diverse AHD environments with different optimization dynamics and feedback patterns.
An AHD environment is naturally specified by three components: the evaluation instances, the solver backbone where the heuristic is used, and the starting point of the design process.
Since evaluation instances can be automatically generated and candidate heuristics can be programmatically evaluated, we can synthesize large-scale AHD RL environments without human annotations.
We therefore construct training data by diversifying these three components.




\textbf{Diversity in seed heuristics, design-set instances, and solver backbones.} We promote generalization by introducing diversity in three aspects of the AHD process. For seed heuristic diversity, we include both seed-guided design, where the model improves seed heuristics collected by ReEvo~\cite{ReEvo:ye2024reevo} with diverse structures and quality levels, and seed-free design, where the model starts only from the task description and interface, with a reference heuristic used only for reward normalization. For design-set instance diversity, we construct design sets with varying instance characteristics, such as instance sizes and node layouts, exposing the model to diverse feedback signals. For solver-backbone diversity, we evaluate heuristics using both constructive and ACO-based solvers, which differ in heuristic interfaces, code structures, and search behaviors. Together, these sources of diversity improve transferable heuristic design.

\subsubsection{Cross-domain RL Training}
\label{subsec:cross-domain-rl}


Existing LLM-based AHD RL methods train the model on a single problem domain~\cite{CALM:huang2025calm,RFTHGS:zhu2026refining}, limiting the model's general AHD ability.
To improve general AHD capability, we jointly train one LLM across multiple problem domains and solver backbones.
Specifically, we use two problem domains and two solver backbones, and construct the corresponding training environments with the synthesis pipeline above.
We train the model with GRPO, with the algorithm and implementation details provided in Appendix C.2.
At test time, the trained model is deployed in the same multi-turn environment, but evaluated on datasets disjoint from those used for training.

\begin{table*}[t]
\centering
\small
\setlength{\tabcolsep}{1.0pt}
\resizebox{\textwidth}{!}{
\begin{tabular}{l*{24}{c}}
\toprule
 & \multicolumn{6}{c}{TSP-Constructive ($\downarrow$)} & \multicolumn{6}{c}{CVRP-Constructive ($\downarrow$)} & \multicolumn{6}{c}{TSP-ACO ($\downarrow$)} & \multicolumn{6}{c}{CVRP-ACO ($\downarrow$)} \\
\cmidrule(lr){2-7} \cmidrule(lr){8-13} \cmidrule(lr){14-19} \cmidrule(lr){20-25}
Method & \multicolumn{2}{c}{N=100} & \multicolumn{2}{c}{N=200} &  &  & \multicolumn{2}{c}{N=100} & \multicolumn{2}{c}{N=200} &  &  & \multicolumn{2}{c}{N=100} & \multicolumn{2}{c}{N=200} &  &  & \multicolumn{2}{c}{N=100} & \multicolumn{2}{c}{N=200} &  &  \\
\cmidrule(lr){2-3} \cmidrule(lr){4-5} \cmidrule(lr){8-9} \cmidrule(lr){10-11} \cmidrule(lr){14-15} \cmidrule(lr){16-17} \cmidrule(lr){20-21} \cmidrule(lr){22-23}
 & Cost & Gap & Cost & Gap & Eval & Cost (\$) & Cost & Gap & Cost & Gap & Eval & Cost (\$) & Cost & Gap & Cost & Gap & Eval & Cost (\$) & Cost & Gap & Cost & Gap & Eval & Cost (\$) \\
\midrule
Optimal & 7.73 & 0.00\% & 10.70 & 0.00\% &  &  & 17.75 & 0.00\% & 32.30 & 0.00\% &  &  & 7.78 & 0.00\% & 10.68 & 0.00\% &  &  & 12.71 & 0.00\% & 22.35 & 0.00\% &  &  \\
Baseline heuristic & 9.59 & 24.09\% & 13.40 & 25.24\% &  &  & 24.28 & 37.28\% & 42.47 & 32.04\% &  &  & 10.08 & 29.50\% & 15.30 & 43.28\% &  &  & 29.96 & 135.9\% & 48.39 & 116.6\% &  &  \\
POMO & 7.94 & 2.79\% & 12.87 & 20.39\% &  &  & 18.44 & 3.84\% & 35.53 & 10.15\% &  &  & 8.01 & 2.95\% & 12.84 & 20.22\% &  &  & 13.30 & 4.64\% & 24.52 & 9.69\% &  &  \\
\midrule
\multicolumn{25}{c}{\textbf{LLM-based AHD: GPT-4o}} \\
\midrule
ReEvo & 9.58 & 24.05\% & 13.38 & 25.06\% & 100 & 0.392 & 22.92 & 29.51\% & 40.52 & 25.88\% & 100 & 0.642 & 8.67 & 11.36\% & 12.81 & 19.94\% & 100 & 0.473 & 32.26 & 147.92\% & 52.87 & 209.5\% & 100 & 1.101 \\
EOH & 9.13 & 18.14\% & 12.75 & 19.24\% & 100 & 0.214 & 23.35 & 31.94\% & 40.99 & 27.34\% & 100 & 0.332 & 8.41 & 8.06\% & 12.22 & 14.39\% & 100 & 0.236 & \underline{15.87} & \underline{25.00\%} & \underline{28.58} & \underline{27.90\%} & 100 & 0.401 \\
MCTS-AHD & 9.27 & 20.00\% & 13.01 & 21.63\% & 100 & 0.073 & 22.59 & 27.58\% & 39.81 & 23.58\% & 100 & 0.702 & 8.37 & 8.54\% & 12.30 & 15.27\% & 100 & 1.026 & 17.80 & 39.88\% & 32.29 & 44.52\% & 100 & 0.986 \\
\methodname{} & 9.56 & 23.73\% & 13.40 & 25.28\% & \textbf{4.6} & 0.123 & 23.62 & 33.47\% & 41.26 & 28.09\% & \textbf{12} & 0.114 & 8.89 & 14.85\% & 13.30 & 24.31\% & \textbf{7} & 0.058 & 27.24 & 114.49\% & 44.78 & 100.39\% & \textbf{7.8} & 0.086 \\
\midrule
\multicolumn{25}{c}{\textbf{LLM-based AHD: DeepSeek-V4-Flash}} \\
\midrule
ReEvo & 9.09 & 17.70\% & 12.84 & 20.07\% & 100 & 0.059 & 22.43 & 26.61\% & 39.66 & 23.05\% & 100 & 0.037 & 8.35 & 7.26\% & 12.23 & 14.51\% & 100 & 0.051 & 17.31 & 36.23\% & 31.74 & 42.00\% & 100 & 0.065 \\
EOH & 9.48 & 22.76\% & 13.29 & 24.23\% & 100 & \underline{0.027} & 22.67 & 28.01\% & 40.22 & 24.75\% & 100 & \underline{0.016} & 8.47 & 8.84\% & 12.37 & 15.80\% & 100 & \underline{0.013} & 16.28 & 28.11\% & 29.20 & 30.66\% & 100 & \underline{0.026} \\
MCTS-AHD & 9.35 & 20.94\% & 13.11 & 22.52\% & 100 & 0.042 & 22.68 & 28.07\% & 40.14 & 24.61\% & 100 & 0.074 & 8.48 & 8.86\% & 12.40 & 16.04\% & 100 & 0.104 & 16.46 & 29.51\% & 29.20 & 30.66\% & 100 & 0.464 \\
\methodname{} & 9.45 & 22.29\% & 13.11 & 22.60\% & 18.4 & 0.054 & 22.08 & 24.60\% & 39.12 & 21.32\% & 20.4 & 0.067 & 8.39 & 7.77\% & 12.12 & 13.42\% & 14.2 & 0.061 & 18.33 & 44.25\% & 32.84 & 46.95\% & 17.6 & 0.066 \\
\midrule
\multicolumn{25}{c}{\textbf{RL Model: Qwen3-4B-Instruct-2507}} \\
\midrule
CALM & \underline{8.85} & \underline{14.60\%} & 13.16 & 23.05\% & 150 & 0.033 & 24.13 & 36.30\% & 43.24 & 34.59\% & 150 & 0.094 & 8.61 & 10.56\% & 12.67 & 18.63\% & 150 & 0.069 & 16.61 & 30.72\% & 30.24 & 35.30\% & 150 & 0.075 \\
\rowcolor{gray!15} \methodname{} & 9.04 & 17.07\% & \underline{12.72} & \underline{18.96\%} & 12.8 & \textbf{0.010} & \underline{21.47} & \underline{21.17\%} & \underline{38.17} & \underline{18.37\%} & \underline{14.6} & \textbf{0.004} & \underline{8.35} & \underline{7.20\%} & \underline{12.04} & \underline{12.71\%} & \underline{11.8} & \textbf{0.003} & 16.60 & 30.67\% & 30.52 & 36.54\% & \underline{12.9} & \textbf{0.004} \\
\rowcolor{gray!15} \methodname{} w/SR & \textbf{8.80} & \textbf{13.86\%} & \textbf{12.31} & \textbf{15.14\%} & 100 & 0.043 & \textbf{21.44} & \textbf{21.03\%} & \textbf{38.16} & \textbf{18.33\%} & 100 & 0.033 & \textbf{8.34} & \textbf{7.12\%} & \textbf{12.02} & \textbf{12.48\%} & 100 & 0.031 & \textbf{15.63} & \textbf{23.15\%} & \textbf{28.10} & \textbf{25.64\%} & 100 & 0.039 \\
\bottomrule
\end{tabular}
}
\vspace{-2mm}
\caption{\textbf{In-domain performance and search efficiency.}
Validation results on the four RL training domains: TSP-Constructive, CVRP-Constructive, TSP-ACO, and CVRP-ACO. Each problem size reports the mean objective and mean Gap (\%).
Eval denotes design-time evaluator calls, and Cost denotes the average USD cost per run.
Method results, Eval, and Cost are averaged over five independent runs.
Reference objectives and seed heuristics are detailed in Appendix D.2.}
\vspace{-2mm}
\label{tab:table20-rl-training-mean-gap-no-para}
\end{table*}

\begin{table*}[t]
\centering
\small
\setlength{\tabcolsep}{0.85pt}
\resizebox{\textwidth}{!}{
\begin{tabular}{l*{24}{c}}
\toprule
 & \multicolumn{8}{c}{OP-ACO ($\uparrow$)} & \multicolumn{8}{c}{MKP-ACO ($\uparrow$)} & \multicolumn{8}{c}{BPP-ACO ($\downarrow$)} \\
\cmidrule(lr){2-9} \cmidrule(lr){10-17} \cmidrule(lr){18-25}
Method & \multicolumn{2}{c}{N=50} & \multicolumn{2}{c}{N=100} & \multicolumn{2}{c}{N=200} &  &  & \multicolumn{2}{c}{N=100} & \multicolumn{2}{c}{N=200} & \multicolumn{2}{c}{N=300} &  &  & \multicolumn{2}{c}{N=500} & \multicolumn{2}{c}{N=1000} & \multicolumn{2}{c}{N=2000} &  &  \\
\cmidrule(lr){2-3} \cmidrule(lr){4-5} \cmidrule(lr){6-7} \cmidrule(lr){10-11} \cmidrule(lr){12-13} \cmidrule(lr){14-15} \cmidrule(lr){18-19} \cmidrule(lr){20-21} \cmidrule(lr){22-23}
 & Cost & Gap & Cost & Gap & Cost & Gap & Eval & Cost (\$) & Cost & Gap & Cost & Gap & Cost & Gap & Eval & Cost (\$) & Cost & Gap & Cost & Gap & Cost & Gap & Eval & Cost (\$) \\
 \midrule
Optimal & 16.02 & 0.00\% & 33.23 & 0.00\% & 62.60 & 0.00\% &  &  & 17.16 & 0.00\% & 44.51 & 0.00\% & 57.01 & 0.00\% &  &  & 200.66 & 0.00\% & 400.41 & 0.00\% & 801.70 & 0.00\% &  &  \\
Baseline heuristic & 13.46 & 16.02\% & 24.38 & 26.64\% & 36.68 & 41.39\% &  &  & 16.50 & 3.69\% & 41.07 & 7.69\% & 52.78 & 8.05\% &  &  & 208.87 & 4.09\% & 417.87 & 4.36\% & 838.15 & 4.55\% &  &  \\
DeepACO & 15.07 & 6.07\% & 30.44 & 8.49\% & 53.34 & 14.71\% &  &  & 16.97 & 1.40\% & 43.51 & 2.11\% & 55.83 & 2.59\% &  &  & 203.07 & 1.21\% & 405.13 & 1.18\% & 810.76 & 1.13\% &  &  \\
\midrule
\multicolumn{25}{c}{\textbf{LLM-based AHD: GPT-4o}} \\
\midrule
ReEvo & 15.04 & 6.14\% & 29.31 & 11.82\% & 51.82 & 17.18\% & 100 & 0.903 & 16.95 & 1.23\% & 42.59 & 4.30\% & 52.48 & 7.07\% & 100 & 0.637 & 206.38 & 2.85\% & 412.48 & 3.01\% & 826.67 & 3.11\% & 100 & 0.592 \\
EOH & 15.08 & 5.89\% & 29.65 & 10.79\% & 52.30 & 16.44\% & 100 & 0.343 & 16.87 & 1.46\% & 42.67 & 3.92\% & 55.13 & 2.88\% & 100 & 0.227 & 204.56 & 1.95\% & 408.53 & 2.03\% & 818.11 & 2.05\% & 100 & 0.277 \\
MCTS-AHD & 14.98 & 6.36\% & 29.86 & 9.90\% & 52.89 & 15.23\% & 100 & 0.930 & 16.92 & 1.44\% & 42.81 & 3.38\% & 52.66 & 2.86\% & 100 & 0.844 & 204.85 & 2.09\% & 409.22 & 2.20\% & 819.24 & 2.19\% & 100 & 0.932 \\
\methodname{} & 13.73 & 14.34\% & 25.00 & 24.79\% & 38.60 & 38.30\% & \textbf{7} & 0.089 & 16.94 & 1.17\% & 43.20 & 2.91\% & 55.67 & 2.18\% & \textbf{5.8} & 0.058 & 210.14 & 4.73\% & 422.37 & 5.48\% & 852.33 & 6.31\% & 8.0 & 0.097 \\
\midrule
\multicolumn{25}{c}{\textbf{LLM-based AHD: DeepSeek-V4-Flash}} \\
\midrule
ReEvo & 14.95 & 6.69\% & 29.12 & 12.37\% & 51.16 & 18.26\% & 100 & 0.047 & 16.96 & 1.25\% & 43.14 & 3.01\% & 55.78 & 1.85\% & 100 & 0.033 & 204.93 & 2.13\% & 409.37 & 2.24\% & 819.64 & 2.24\% & 100 & 0.087 \\
EOH & 15.06 & 6.07\% & 30.10 & 9.45\% & 53.82 & 14.02\% & 100 & \underline{0.018} & 17.01 & 0.92\% & 43.32 & 2.61\% & 55.81 & 1.82\% & 100 & \underline{0.011} & 204.49 & 1.91\% & 408.44 & 2.01\% & 817.48 & 1.97\% & 100 & 0.038 \\
MCTS-AHD & 15.14 & 5.53\% & 30.27 & 8.93\% & 54.30 & 13.25\% & 100 & 0.074 & 16.94 & 1.31\% & 42.65 & 4.13\% & 55.57 & 2.45\% & 100 & 0.050 & \underline{204.32} & \underline{1.83\%} & \underline{407.97} & \underline{1.89\%} & \textbf{816.45} & \textbf{1.84\%} & 100 & 0.097 \\
\methodname{} & 14.95 & 6.74\% & 29.21 & 12.10\% & 50.78 & 18.84\% & 19.0 & 0.065 & 17.02 & 0.86\% & 43.30 & 2.66\% & 55.79 & 1.91\% & 15.2 & 0.040 & 206.32 & 2.82\% & 412.53 & 3.03\% & 826.71 & 3.12\% & 26.8 & \underline{0.017} \\
\midrule
\multicolumn{25}{c}{\textbf{RL Model: Qwen3-4B-Instruct-2507}} \\
\midrule
CALM & 14.99 & 6.47\% & 29.81 & 10.29\% & 53.18 & 15.04\% & 150 & 0.074 & 16.91 & 1.52\% & 42.91 & 3.52\% & 55.41 & 2.43\% & 150 & 0.053 & 209.25 & 4.28\% & 419.31 & 4.72\% & 842.50 & 5.09\% & 150 & 0.042 \\
\rowcolor{gray!15}\methodname{} & \underline{15.17} & \underline{5.34\%} & \underline{30.36} & \underline{8.62\%} & \underline{54.56} & \underline{12.84\%} & \underline{13.6} & \textbf{0.005} & \underline{17.03} & \underline{0.84\%} & \underline{43.52} & \underline{2.17\%} & \underline{56.19} & \underline{1.19\%} & \underline{12.4} & \textbf{0.004} & 206.05 & 2.69\% & 411.62 & 2.80\% & 824.26 & 2.81\% & \underline{11.7} & \textbf{0.007} \\
\rowcolor{gray!15}\methodname{} w/SR & \textbf{15.36} & \textbf{4.18\%} & \textbf{30.97} & \textbf{6.60\%} & \textbf{55.84} & \textbf{10.47\%} & 100 & 0.034 & \textbf{17.06} & \textbf{0.67\%} & \textbf{43.84} & \textbf{1.44\%} & \textbf{56.57} & \textbf{0.59\%} & 100 & 0.033 & \textbf{204.21} & \textbf{1.77\%} & \textbf{407.93} & \textbf{1.88\%} & \underline{816.94} & \underline{1.90\%} & 100 & 0.044 \\
\bottomrule
\end{tabular}
}
\vspace{-2mm}
\caption{\textbf{Generalization to unseen combinatorial domains.}
Validation results on OP-ACO, MKP-ACO, and BPP-ACO, which are excluded from RL training.
Method results, Eval, and Cost are averaged over five independent runs.
Reference objectives and seed heuristics are detailed in Appendix D.2.}
\vspace{-6mm}
\label{tab:table21-generalization-mean-gap-no-para}
\end{table*}

\textbf{Reward design.}
\label{subsec:rewardDesign}
We use the score of the final heuristic as the reward signal for the whole trajectory.
If no valid code can be extracted, the trajectory receives a fixed penalty.
If the code fails during execution or produces an infeasible solution, it receives another fixed penalty.
For a feasible final heuristic \(h_{\mathrm{final}}\), we measure its improvement over a baseline heuristic \(h_b\) on the design set, defined as \(\Delta(h_{\mathrm{final}})=\text{Score}(h_{\mathrm{final}};\mathcal{D}_{\mathrm{design}})-\text{Score}(h_b;\mathcal{D}_{\mathrm{design}})\).
Here \(h_b\) is the provided seed heuristic for seed-guided tasks, and a default human-designed heuristic for seed-free tasks.
The reward is
\[
\mathcal{R} =
\begin{cases}
-2.0, & \text{no code is extracted},\\
-1.5, & \text{execution failure or infeasible solution},\\
\Delta(h_{\mathrm{final}}), & \text{feasible heuristic}.
\end{cases}
\]
This design discourages invalid code while directly rewarding improvements over the baseline associated with each synthesized environment.

\section{Experiments}

\subsection{Experimental Setup}

\textbf{Problems.} We evaluate \methodname{} on combinatorial and continuous heuristic-design benchmarks. The combinatorial suite includes constructive routing tasks (TSP and CVRP)~\citep{TSPProblem:lawler1985traveling,LLM4AD:liu2024llm4ad} and ACO-based tasks (TSP, CVRP, orienteering problem (OP), multiple knapsack problem (MKP), and offline bin packing problem (Offline BPP))~\citep{ACO:dorigo2006ant}, covering routing, knapsack, and packing structures across two solver backbones. We also assess cross-protocol transfer on cost-aware Bayesian optimization (CAF)~\citep{Bo-CAF:yao2024evolve}. Detailed formulations, interfaces, instance-generation protocols, and validation settings are provided in Appendix A.

\noindent \textbf{Baselines.} We compare \methodname{} with problem-specific methods, fixed-workflow LLM-AHD methods, and CALM, which fine-tunes the LLM within a fixed evolutionary workflow~\citep{CALM:huang2025calm}. Problem-specific methods include Greedy-Construct (GC)~\cite{TSPGC:rosenkrantz1977analysis}, standard ACO heuristics~\cite{TSPACO:skinderowicz2022improving,CVRPACO:cai2022dynamic,OPACO:sohrabi2021acs,MKPACO:fidanova2020hybrid}, classical CAF acquisition functions~\citep{Bo-CAF:yao2024evolve}, POMO on TSP/CVRP~\citep{POMO:kwon2020pomo}, and DeepACO on ACO-based tasks~\citep{DeepACO:ye2023deepaco}. Fixed-workflow baselines are ReEvo~\citep{ReEvo:ye2024reevo}, EOH~\citep{EOH:liu2024evolution}, and MCTS-AHD~\citep{MCTS-AHD:zheng2025monte}, evaluated with GPT-4o~\citep{GPT-4o:hurst2024gpt} and DeepSeek-V4-Flash~\citep{Deepseek-V4:deepseekai2026deepseekv4}. All LLM-AHD methods use the same task interfaces, data, objective computation, and seed heuristics.

\begin{table*}[h]
\centering

\resizebox{\textwidth}{!}{
\begin{tabular}{l|cc|cccccccccc|ccc}
\toprule
Method & Ackley & Rastrigin & Griewank & Rosenbrock & Levy & ThreeHumpCamel & StyblinskiTang & Hartmann-3D & Powell & Shekel & Hartmann-6D & Cosine8 & Avg. & Eval & Cost (\$) \\
\midrule
EI & 2.66\% & 4.74\% & 0.49\% & 1.26\% & \textbf{0.01\%} & \textbf{0.05\%} & 0.03\% & \textbf{0.00\%} & 18.89\% & 7.91\% & \textbf{0.03\%} & 0.47\% & 3.05\% &  &  \\
EIpu & \textbf{2.33\%} & 5.62\% & \textbf{0.34\%} & 2.36\% & \textbf{0.01\%} & 0.12\% & \textbf{0.02\%} & \textbf{0.00\%} & 19.83\% & 7.92\% & \textbf{0.03\%} & 0.47\% & 3.25\% &  &  \\
EI-cool & 2.74\% & 5.78\% & \textbf{0.34\%} & 2.29\% & \textbf{0.01\%} & 0.07\% & 0.03\% & \textbf{0.00\%} & 14.95\% & 8.21\% & \textbf{0.03\%} & 0.54\% & 2.92\% &  &  \\
\midrule
\multicolumn{16}{c}{\textbf{LLM-based AHD: DeepSeek-V4-Flash}} \\
\midrule
EOH  & 3.14\% & 4.02\% & \textbf{0.23\%} & 1.63\% & 0.02\% & 0.16\% & 0.18\% & 0.00\% & 32.02\% & 7.38\% & 0.18\% & 0.50\% & 4.12\% & 100 & 0.025 \\
MCTS-AHD  & 3.19\% & 5.29\% & 0.34\% & \textbf{0.83\%} & 0.04\% & 0.07\% & 0.12\% & 0.01\% & 26.36\% & 7.59\% & 0.14\% & 0.61\% & 3.72\% & 100 & 0.071 \\
ReEvo  & 4.82\% & 5.25\% & 1.09\% & 11.02\% & 0.23\% & 0.29\% & 2.13\% & 0.05\% & 73.56\% & 8.70\% & 0.61\% & 0.75\% & 9.04\% & 100 & 0.153 \\
\midrule
\multicolumn{16}{c}{\textbf{RL Model: Qwen3-4B-Instruct-2507}} \\
\midrule
CALM & 3.28\% & 4.22\% & 1.01\% & 23.05\% & 0.31\% & 0.57\% & 1.61\% & 0.04\% & 50.38\% & 8.44\% & 0.60\% & 0.86\% & 7.86\% & 150 & 0.057 \\
\rowcolor{gray!15}\methodname{} & 2.88\% & \textbf{1.73\%} & 0.35\% & 2.99\% & \textbf{0.01\%} & \textbf{0.05\%} & 0.49\% & 0.11\% & \textbf{7.00\%} & \textbf{5.02\%} & 0.47\% & \textbf{0.40\%} & \textbf{1.79\%} & \textbf{14.2} & \textbf{0.007} \\
\bottomrule
\end{tabular}
}
\caption{\textbf{Generalization to cost-aware Bayesian optimization.} CAF comparison on 12 benchmark functions. Ackley and Rastrigin are used as design functions during AHD search, while the remaining ten functions are held-out validation functions. Entries report gaps to the known optimum with BO budget 30 and 10 trials per test function; lower is better. The results of EI, EIpu, and EI-cool are from~\cite{MCTS-AHD:zheng2025monte}. Method results, Eval, and Cost are averaged over three independent runs.}
\label{tab:caf-table4-comparison}
\end{table*}

\begin{figure*}[t]
\centering

\begin{subfigure}[t]{0.49\textwidth}
    \centering
    \includegraphics[width=\linewidth]{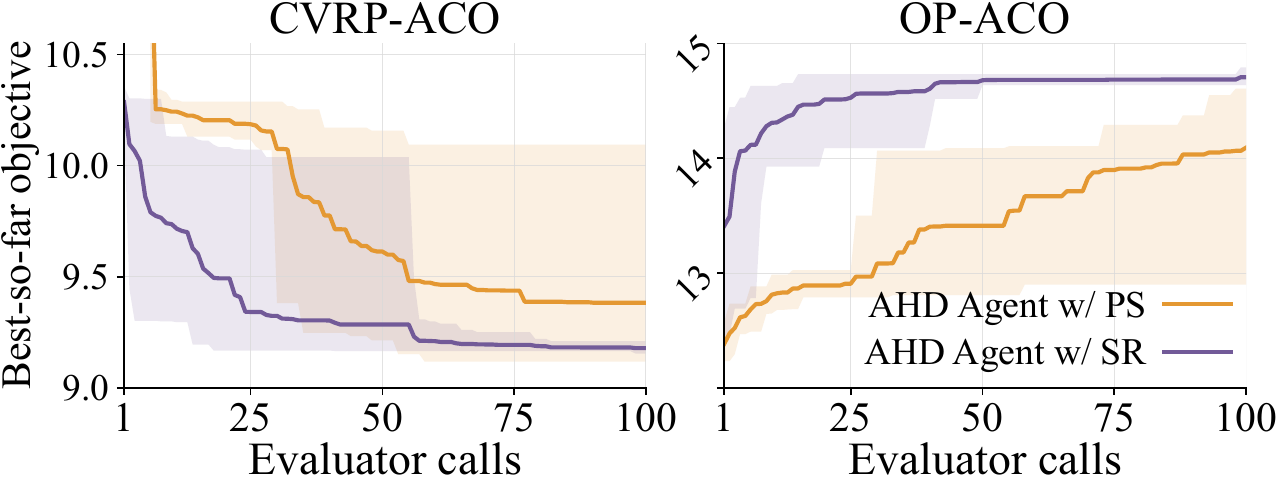}
    \caption{}
    \label{fig:inference-time-scaling-ps-vs-sr}
\end{subfigure}
\hfill
\begin{subfigure}[t]{0.49\textwidth}
    \centering
    \includegraphics[width=\linewidth]{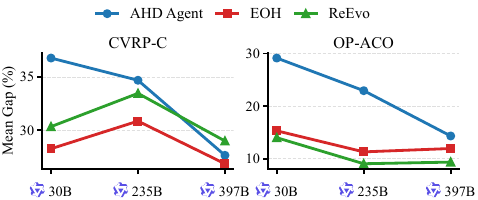}
    \caption{}
    \label{fig:qwen-two-domain-gap-scaling}
\end{subfigure}

\caption{ \textbf{Scaling Effect of \methodname{}.}
(a) \textbf{Inference-time scaling comparison.} SR strategy outperforms PS strategy on two tasks.
(b) \textbf{Model scaling favors \methodname{}.} Performance of \methodname{} increases as the model size increases from 30B to 397B parameters. 
\vspace{-4mm}}
\label{fig:pipe-vs-long-multiturn-focus}

\end{figure*}




\noindent \textbf{RL training.}
We use Qwen3-4B-Instruct-2507~\citep{Qwen3:qwen3technicalreport} with GRPO on 4{,}000 prompts sampled equally from four training domains: TSP-Constructive, CVRP-Constructive, TSP-ACO, and CVRP-ACO.
Generalization is evaluated on three unseen combinatorial domains(OP-ACO, MKP-ACO and BPP-ACO) and the cross-protocol CAF benchmark. All experiments are conducted on 4$\times$A100-80G GPUs with Intel Xeon 8358P CPUs, and total RAM is 1007GB. Detailed information is provided in Appendix C.

\noindent \textbf{Hyperparameter settings.} We use validation Gap~(\%) as the primary metric, where lower is better. Following ReEvo~\citep{ReEvo:ye2024reevo}, fixed-workflow baselines and \methodname{} w/ SR use a budget of 100 heuristic evaluations. The solver and evaluation settings for all baselines follow ReEvo. To further ensure runtime fairness, each heuristic evaluation is capped at 60 seconds during design and 180 seconds during test evaluation. CALM follows its original 150-evaluation setting, while the short \methodname{} setting uses at most 30 evaluations. We also report the mean USD inference cost per run; pricing details are provided in Appendix D.1.

\subsection{Overall Results}
\label{sec:overallResult}

\paragraph{Training-domain efficiency.} As shown in Table~\ref{tab:table20-rl-training-mean-gap-no-para}, the RL-trained \methodname{} is efficient across all four training domains. With only 11--15 heuristic evaluations, the short setting remains competitive with baselines using 100 evaluations, while \methodname{} w/ SR achieves the lowest gap among the compared LLM-AHD methods on all four domains under the 100-evaluation budget. The DeepSeek-V4-Flash variant also remains competitive on TSP-ACO and CVRP-Constructive with a small evaluation budget, suggesting that \methodname{} remains highly efficient when paired with stronger general-purpose LLMs. Statistical significance results are reported in Appendix D.9.

\noindent \textbf{Transfer across held-out combinatorial domains.} Table~\ref{tab:table21-generalization-mean-gap-no-para} evaluates three domains excluded from RL training: OP-ACO, MKP-ACO, and BPP-ACO. While OP remains related to the routing tasks seen during training, MKP and BPP introduce non-routing objectives and substantially different constraint structures based on item selection and packing. With only 12--14 heuristic evaluations, the short \methodname{} setting remains competitive on OP and MKP and improves over vanilla ACO on BPP. At 100 evaluations, \methodname{} w/ SR outperforms the problem-specific DeepACO baseline at every tested size on both OP and MKP, demonstrating transfer beyond routing-oriented training domains. On BPP, it substantially narrows the gap to DeepACO but remains weaker, revealing a clear boundary of the learned policy.

\noindent\textbf{Transfer to continuous optimization.} As shown in Table~\ref{tab:caf-table4-comparison}, despite being trained only on combinatorial tasks, \methodname{} transfers to the continuous CAF benchmark. It achieves an average gap of 1.79\% over 12 Bayesian-optimization functions with only 14.2 heuristic evaluations, outperforming hand-crafted acquisition baselines, such as EI-cool (2.92\%) and the DeepSeek-V4-Flash-based EOH (4.12\%), MCTS-AHD (3.72\%), and ReEvo (9.04\%). These results show that the learned agentic design policy can transfer across evaluation protocols, not only across combinatorial problem families.

\subsection{Further Experiments}


\textbf{Inference-time scaling.} Figure~\ref{fig:inference-time-scaling-ps-vs-sr} compares two budget-scaling strategies: \methodname{} w/ SR extends one session to exploit accumulated feedback, whereas \methodname{} w/ PS independently samples multiple short trajectories and selects the best candidate. Both improve with larger budgets, but continuing a coherent trajectory is generally more effective. Parallel sampling remains useful under wall-clock constraints because its trajectories can run concurrently.

\noindent\textbf{Model scaling.}
Figure~\ref{fig:qwen-two-domain-gap-scaling} shows that the agentic multi-turn paradigm benefits more consistently from stronger LLM backbones than fixed-workflow AHD methods. On OP-ACO, the \methodname{} gap decreases from 29.14\% with the 30B Qwen model to 14.35\% with the 397B model. In contrast, EOH and ReEvo show non-monotonic or domain-dependent trends under the same backbone sweep. This suggests that stronger reasoning capability is more effectively converted into performance when the LLM controls the design process, including feedback interpretation, tool use, and iterative revision, rather than only generating candidates inside a fixed workflow.

\subsection{Training Curves}

\begin{figure}[t]
\centering
\includegraphics[width=\linewidth]{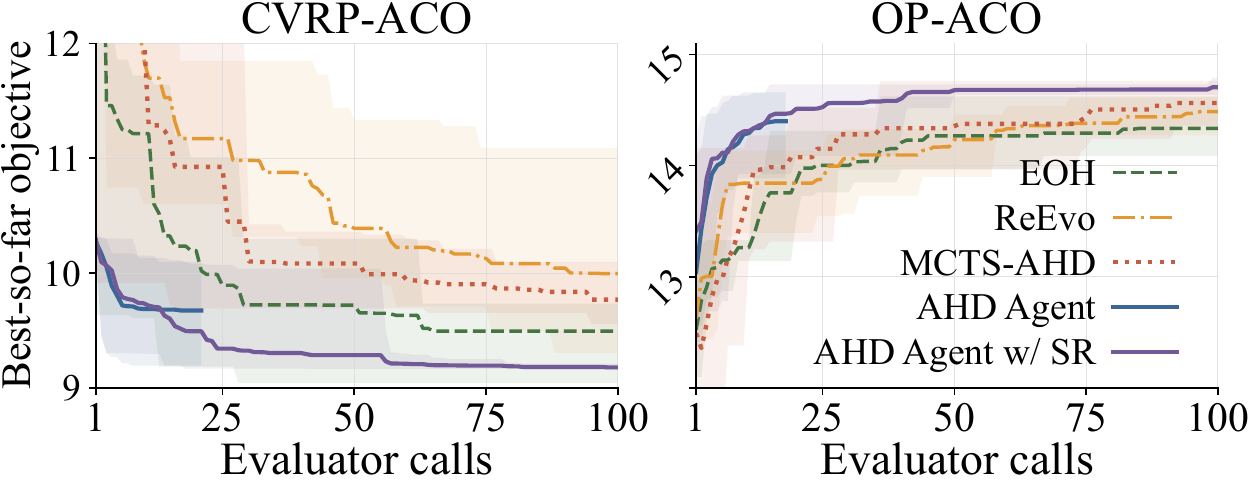}
\caption{
\textbf{Training curves during design.}
\methodname{} converges faster and achieves better performance under larger evaluation budgets.
}
\label{fig:train_convergence}
\vspace{-2mm}
\end{figure}

\begin{figure}[t]
\centering
\includegraphics[width=\linewidth]{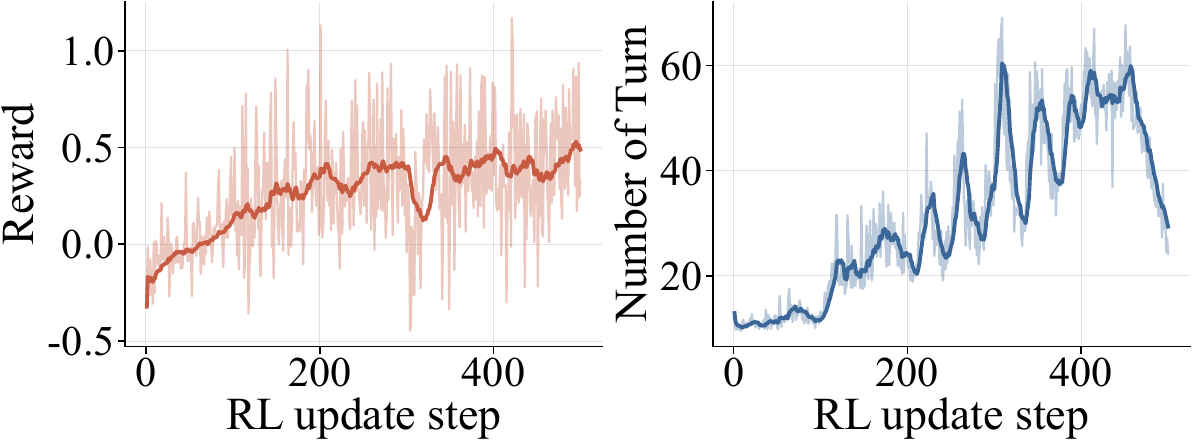}
\caption{\textbf{RL training dynamics.} Reward and the number of turns increase after a 500-step RL training.}
\label{fig:rl-training-reward-turns}
\vspace{-3mm}
\end{figure}

\textbf{Training curves during design.}
We further visualize the training curves during the design process. After RL training, we use a separate design set, distinct from the training set used in RL training, to obtain the heuristics for testing. For plotting, we align the five independent design runs by evaluator-call index and use the longest run as the curve horizon; runs that stop earlier are padded by carrying forward their final best-so-far gap.
\cref{fig:train_convergence} shows that our method converges faster than the baselines while requiring significantly fewer evaluations.
Moreover, when scaling the evaluation budget with SR, our method converges to substantially better performance.

\noindent \textbf{RL training dynamics.} Figure~\ref{fig:rl-training-reward-turns} shows that the reward steadily increases over 500 training steps, indicating progressively better heuristic revisions. The number of turns initially increases as the agent shifts from one-shot generation to feedback-driven decisions over generation, evaluation, and tool use, and then stabilizes at around 30, suggesting that the policy does not simply prolong the interaction. Further details are provided in Appendix C.3.

\subsection{Ablation Study}

\begin{figure}[t]
\centering
\includegraphics[width=0.7\linewidth]{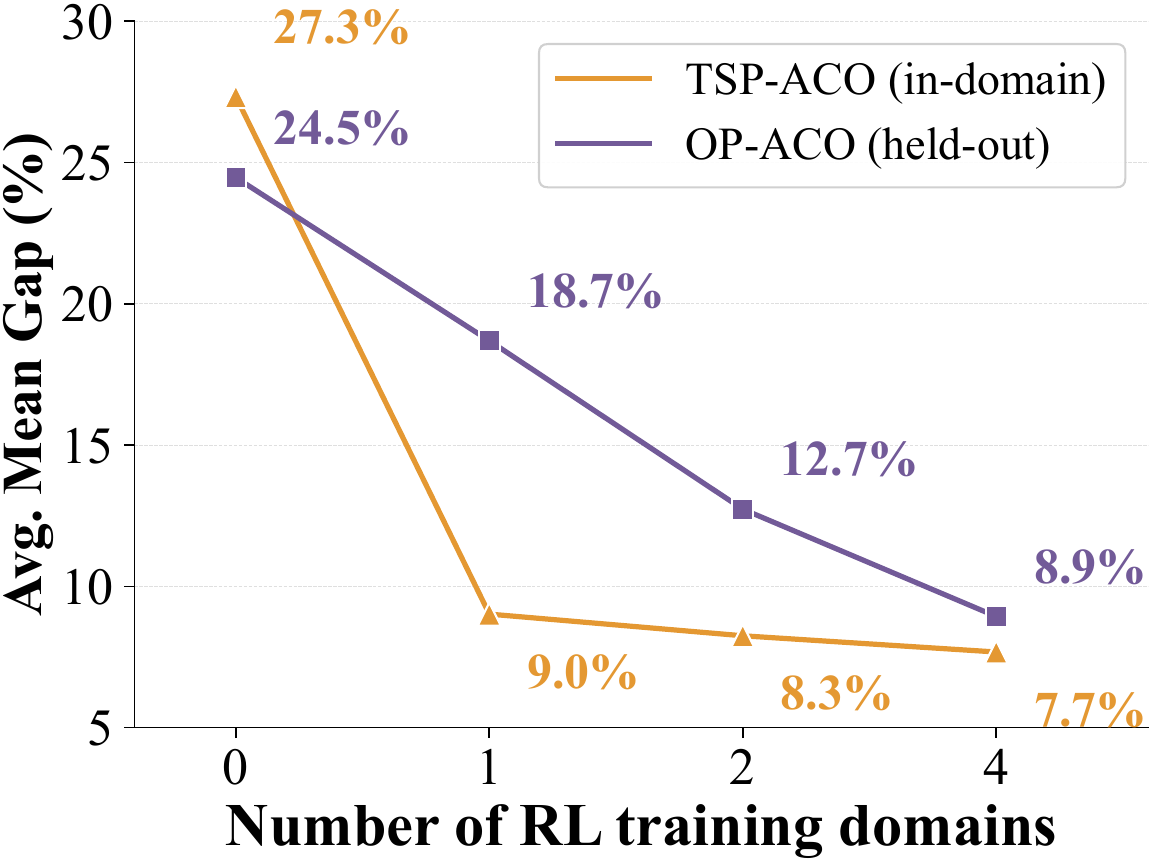}
\caption{Performance changes as the number of training domains increases.}
\label{fig:cross-domain-slopes}
\end{figure}

\textbf{Cross-domain RL training increases the general AHD capability.}
\label{sec:cross-domain-ablation}
Figure~\ref{fig:cross-domain-slopes} shows how the performance on an in-domain task and a held-out task changes as the number of RL training domains increases.
The training mixture is gradually expanded by adding TSP-ACO, CVRP-ACO, and TSP/CVRP-Constructive domains.
Note that all runs are trained for the same 500 steps.
On OP-ACO, which is never included in the RL training mixture, the gap decreases steadily from \(24.5\%\) to \(8.9\%\) as more training domains are added.
This consistently held-out improvement indicates that cross-domain RL training learns transferable heuristic-design behavior rather than fitting domain-specific heuristic templates.
Meanwhile, on in-domain TSP-ACO, RL reduces the gap from \(27.3\%\) to \(9.0\%\), and adding more domains further improves it to \(7.7\%\). This suggests that broader training mixtures improve both held-out generalization and in-domain AHD performance.

\begin{table}[t]
\centering
\scriptsize
\setlength{\tabcolsep}{3.2pt}
\caption{\textbf{Ablation study of tool-access and RL.} Results evaluated on CVRP-C and CVRP-ACO. Entries are average Mean/Best Gap~(\%) over $N=50,100,200$.}
\label{tab:rl-tool-ablation-cvrp}
\resizebox{\linewidth}{!}{
\begin{tabular}{lcccc}
\toprule
Setting & \multicolumn{2}{c}{CVRP-C $\downarrow$}  & \multicolumn{2}{c}{CVRP-ACO $\downarrow$}  \\
\cmidrule(lr){2-3} \cmidrule(lr){4-5}
& Mean Gap (\%) & Best Gap (\%) & Mean Gap (\%) & Best Gap (\%) \\
\midrule
RL + Full Tools & \textbf{21.197} & \textbf{20.926} & \textbf{29.998} & \textbf{21.999} \\
\midrule
RL + Eval Only & 22.895 & \textbf{20.926} & 33.212 & 22.339 \\
Without RL & 37.587 & 37.411 & 125.806 & 108.397 \\
\bottomrule
\end{tabular}
}
\end{table}

\noindent \textbf{Effectiveness of RL training and tool access.}
Table~\ref{tab:rl-tool-ablation-cvrp} shows that RL training is the main source of improvement, while diagnostic tools provide additional gains. Removing RL while keeping the same multi-turn framework increases the mean gap from 21.20\% to 37.59\% on CVRP-Constructive, and from 30.00\% to 125.81\% on CVRP-ACO. Replacing the full diagnostic tool library with the evaluator-only setting also degrades performance, increasing the mean gap from 21.20\% to 22.90\% on CVRP-Constructive and from 30.00\% to 33.21\% on CVRP-ACO. These experimental results validate the effectiveness of our design.

\section{Conclusion}

We introduced \methodname{}, an agentic reinforcement-learning framework for automatic heuristic design. By letting the LLM Agent control a multi-turn design process and by specializing a compact base model with GRPO on cross-domain heuristic-design tasks, \methodname{} improves search efficiency while reducing reliance on large general-purpose LLMs. Experiments across combinatorial and continuous optimization benchmarks show that the learned policy can effectively revise heuristics with limited evaluator feedback, generalize to unseen domains, and transfer across evaluation protocols. These results suggest that agentic, RL-trained LLMs are a promising alternative to fixed-workflow LLM-AHD methods. 

\bibliography{aaai2027}

\clearpage
\setcounter{secnumdepth}{2}
\appendix


\section{Detail of Problem Domains}
\label{appendix:detail-problem}
\subsection{Problem Domain Definitions}
\label{appendix:problem-domains}

The main paper evaluates eight problem domains spanning combinatorial and
continuous optimization: four RL training domains (TSP-Constructive,
CVRP-Constructive, TSP-ACO, and CVRP-ACO) and four held-out domains (OP-ACO,
MKP-ACO, BPP-ACO, and CAF). Each entry below states the mathematical
formulation, the design/validation instance sizes, and the function interface
that the LLM is asked to design. In addition to the eight domains evaluated in
the main paper, we report supplementary results on OVRP-Constructive; OVRP is
not counted as one of the eight main-paper domains.

\subsubsection{TSP-Constructive}

The Traveling Salesman Problem (TSP) asks for a minimum-cost Hamiltonian cycle over
$n$ cities. Given a distance matrix $D\in\mathbb{R}^{n\times n}$, the objective is
\[
\min_{\sigma\in S_n}\sum_{i=1}^{n}D_{\sigma(i),\sigma(i+1)},
\]
where $\sigma(n+1)\equiv\sigma(1)$. In the constructive setting, the LLM designs a
greedy selection rule that builds a tour one city at a time.

\textbf{Function interface.}
\begin{tcolorbox}[colback=gray!3, colframe=gray!50, left=3pt, right=3pt, top=2pt, bottom=2pt, boxrule=0.4pt]
\small\ttfamily
def select\_next\_node(current\_node, destination\_node, unvisited\_nodes, distance\_matrix):\\
\hspace*{1em}"""Return the index of the next city to visit."""\\
\hspace*{1em}return next\_node \hfill\textrm{\scriptsize\sffamily\color{gray}$\triangleright$ int}
\end{tcolorbox}

\textbf{Instance sizes.} Design set: $N=50$; validation: $N=50, 100, 200$.

\subsubsection{CVRP-Constructive}

The Capacitated Vehicle Routing Problem (CVRP) extends TSP with a depot node and
per-customer demands $d_i$. A fleet of homogeneous vehicles, each with capacity $Q$,
must serve all customers while minimizing total travel distance:
\[
\min \sum_{k}\sum_{(i,j)\in R_k} D_{i,j},
\quad\text{s.t.}\quad
\sum_{i\in R_k}d_i\le Q\;\;\forall k,
\]
where $R_k$ denotes the route of vehicle $k$. In the constructive setting, the LLM
designs the next-customer selection rule, which must also decide when to return to the
depot and start a new route.

\textbf{Function interface.}
\begin{tcolorbox}[colback=gray!3, colframe=gray!50, left=3pt, right=3pt, top=2pt, bottom=2pt, boxrule=0.4pt]
\small\ttfamily
def select\_next\_node(current\_node, depot, feasible\_unvisited,\\
\hspace*{6em}capacity\_remaining, demands, distance\_matrix):\\
\hspace*{1em}"""Return the next customer index, or 0 (depot) to start a new route."""\\
\hspace*{1em}return next\_node \hfill\textrm{\scriptsize\sffamily\color{gray}$\triangleright$ int}
\end{tcolorbox}

\textbf{Instance sizes.} Design set: $N=50$; validation: $N=50, 100, 200$.

\subsubsection{TSP-ACO}

In the ACO setting for TSP, a colony of ants constructs tours in parallel by sampling
transitions with probabilities proportional to $\tau_{ij}^\alpha\cdot\eta_{ij}^\beta$,
where $\tau$ is the pheromone matrix and $\eta$ is the heuristic desirability matrix.
The LLM designs the function that computes $\eta$ from the distance matrix.

\textbf{Function interface.}
\begin{tcolorbox}[colback=gray!3, colframe=gray!50, left=3pt, right=3pt, top=2pt, bottom=2pt, boxrule=0.4pt]
\small\ttfamily
def heuristic(distance\_matrix: np.ndarray):\\
\hspace*{1em}"""Return an (n, n) heuristic desirability matrix."""\\
\hspace*{1em}return heuristic\_matrix \hfill\textrm{\scriptsize\sffamily\color{gray}$\triangleright$ np.ndarray}
\end{tcolorbox}

\textbf{Instance sizes.} Design set: $N=50$; validation: $N=50, 100, 200$.

\subsubsection{CVRP-ACO}

The ACO formulation of CVRP follows the same pheromone-driven construction as TSP-ACO
but incorporates capacity constraints. The LLM designs the heuristic matrix that guides
ant transitions, taking into account both distances and remaining vehicle capacity.

\textbf{Function interface.}
\begin{tcolorbox}[colback=gray!3, colframe=gray!50, left=3pt, right=3pt, top=2pt, bottom=2pt, boxrule=0.4pt]
\small\ttfamily
def heuristic(distance\_matrix: np.ndarray, coordinates: np.ndarray,\\
\hspace*{5.5em}demands: np.ndarray, capacity: float):\\
\hspace*{1em}"""Return an (n, n) heuristic desirability matrix for capacitated routing."""\\
\hspace*{1em}return heuristic\_matrix \hfill\textrm{\scriptsize\sffamily\color{gray}$\triangleright$ np.ndarray}
\end{tcolorbox}

\textbf{Instance sizes.} Design set: $N=50$; validation: $N=50, 100, 200$.

\subsubsection{OP-ACO}

The Orienteering Problem (OP) maximizes the total prize collected by visiting a subset
of nodes, subject to a maximum route length $L$:
\[
\max_{\sigma}\sum_{i\in\sigma}p_i,
\quad\text{s.t.}\quad
\sum_{i}D_{\sigma(i),\sigma(i+1)}\le L.
\]
The ACO framework is used with the LLM designing the heuristic matrix that balances
prize attractiveness against distance cost.

\textbf{Function interface.}
\begin{tcolorbox}[colback=gray!3, colframe=gray!50, left=3pt, right=3pt, top=2pt, bottom=2pt, boxrule=0.4pt]
\small\ttfamily
def heuristic(prize: np.ndarray, distance: np.ndarray, maxlen: float):\\
\hspace*{1em}"""Return an (n, n) heuristic desirability matrix for prize collection."""\\
\hspace*{1em}return heuristic\_matrix \hfill\textrm{\scriptsize\sffamily\color{gray}$\triangleright$ np.ndarray}
\end{tcolorbox}

\textbf{Instance sizes.} RL training: not used (held-out domain). Design set: $N=50$; validation: $N=50, 100, 200$.

\subsubsection{MKP-ACO}

The Multidimensional Knapsack Problem (MKP) maximizes the total value of selected items
subject to multiple resource constraints:
\[
\max_{x\in\{0,1\}^n}\sum_{j=1}^{n}v_j x_j,
\quad\text{s.t.}\quad
\sum_{j=1}^{n}w_{ij}x_j\le C_i,\;\;i=1,\ldots,m.
\]
The ACO framework constructs solutions by sequentially adding items. The LLM designs
the heuristic desirability function that guides item selection.

\textbf{Function interface.}
\begin{tcolorbox}[colback=gray!3, colframe=gray!50, left=3pt, right=3pt, top=2pt, bottom=2pt, boxrule=0.4pt]
\small\ttfamily
def heuristic(prize: np.ndarray, weight: np.ndarray):\\
\hspace*{1em}"""Return an (n,) heuristic desirability vector for item selection."""\\
\hspace*{1em}return heuristic\_vector \hfill\textrm{\scriptsize\sffamily\color{gray}$\triangleright$ np.ndarray}
\end{tcolorbox}

\textbf{Instance sizes.} RL training: not used (held-out domain). Design set: $N=100$; validation: $N=100, 200, 300$.

\subsubsection{Offline BPP-ACO}

The one-dimensional Offline Bin Packing Problem (BPP) assigns a known set of
items with integer sizes $s_i$ to identical bins of capacity $C$ while
minimizing the number of used bins:
\[
\begin{aligned}
\min_{x,y}\quad & \sum_b y_b, \\
\text{s.t.}\quad & \sum_b x_{ib}=1 && \forall i, \\
& \sum_i s_i x_{ib}\le C y_b && \forall b,
\end{aligned}
\]
where $x_{ib},y_b\in\{0,1\}$. The complete item set is available before
packing, so this is an offline rather than online protocol. In the ACO host,
each ant constructs a packing by selecting items for the current bin. The LLM
designs an item-pair desirability matrix whose entry $\eta_{ij}$ estimates how
promising it is to place items $i$ and $j$ in the same bin.

\textbf{Function interface.}
\begin{tcolorbox}[colback=gray!3, colframe=gray!50, left=3pt, right=3pt, top=2pt, bottom=2pt, boxrule=0.4pt]
\small\ttfamily
def heuristics(demand: np.ndarray, capacity: int):\\
\hspace*{1em}"""Return an (n, n) item-pair desirability matrix."""\\
\hspace*{1em}return heuristic\_matrix \hfill\textrm{\scriptsize\sffamily\color{gray}$\triangleright$ np.ndarray}
\end{tcolorbox}

\textbf{Instance generation and sizes.} Item sizes are sampled independently
and uniformly from the integers $\{20,\ldots,100\}$ and the bin capacity is
$C=150$, following the offline BPP setting used by the ACO
host~\citep{BPPF:falkenauer1996hybrid}. BPP-ACO is excluded from RL training. The
evaluation-time design set contains 5 instances at $N=500$; held-out validation
uses $N\in\{500,1000,2000\}$ with 64 instances per size.

\subsubsection{CAF (Cost-Aware Bayesian Optimization)}

In cost-aware Bayesian optimization~\citep{Bo-CAF:yao2024evolve}, the goal is to design
a utility (acquisition) function that guides the selection of the next evaluation point.
Given a surrogate model's predictions, the LLM designs a function that maps posterior
statistics and cost information to a scalar utility value.

\textbf{Function interface.}
\begin{tcolorbox}[colback=gray!3, colframe=gray!50, left=3pt, right=3pt, top=2pt, bottom=2pt, boxrule=0.4pt]
\small\ttfamily
def utility(train\_x, train\_Y, best\_x, best\_Y, test\_x,\\
\hspace*{4.5em}mean\_test\_y, std\_test\_y, cost\_test\_y,\\
\hspace*{4.5em}budget\_used, budget\_total):\\
\hspace*{1em}"""Return a scalar acquisition value for each candidate point."""\\
\hspace*{1em}return utility\_value \hfill\textrm{\scriptsize\sffamily\color{gray}$\triangleright$ torch.Tensor}
\end{tcolorbox}

\textbf{Test protocol.} 12 benchmark functions (Ackley, Rastrigin, Griewank, Rosenbrock,
Levy, ThreeHumpCamel, StyblinskiTang, Hartmann-3D, Powell, Shekel, Hartmann-6D, Cosine8),
BO budget 30, 10 trials per instance. Ackley and Rastrigin are training instances, others are val instances.

\subsubsection{Additional Supplementary Domain: OVRP-Constructive}

The Open Vehicle Routing Problem (OVRP) is a variant of CVRP where vehicles are
not required to return to the depot after serving their last customer. The LLM
designs the constructive selection rule under this modified termination
condition.

\textbf{Function interface.} Same as CVRP-Constructive.

\textbf{Instance sizes.} RL training: not used. Design set: $N=50$;
validation: $N=50,100,200$. OVRP-Constructive is reported only as an additional
supplementary domain and is not part of the eight-domain main-paper benchmark.

\subsection{Algorithmic Framework Definitions}
\label{appendix:frameworks}

The LLM-designed heuristic is embedded into one of three algorithmic frameworks.
Each framework defines the overall solving procedure; the LLM is responsible for
designing a specific component within that procedure. Below we describe each
framework in detail, including the pseudocode and the exact role of the LLM-designed
component.

\subsubsection{Constructive Heuristic Framework}

Step-by-step construction is an intuitive and general-purpose framework for combinatorial optimization. It builds a feasible solution incrementally from scratch: starting from an empty partial solution, the framework repeatedly selects the most promising candidate element and appends it, until the solution is complete. This greedy construction paradigm is applicable to a wide range of CO problems, including routing, packing, scheduling, and assignment problems.

The core of the framework is a \emph{selection function} that directly returns
the next feasible element based on the current partial solution state. The
framework handles feasibility checks (e.g., capacity constraints in vehicle
routing), termination conditions, and final objective evaluation. The framework
enforces feasibility for valid function outputs, while invalid outputs or
execution failures are rejected by the evaluator.

For TSP, the classical manually designed baseline is the nearest-neighbor heuristic~\citep{TSPGC:rosenkrantz1977analysis}, which selects unvisited cities solely by their Euclidean distance from the current city. This simple rule produces tours that are typically 20--25\% longer than optimal. For CVRP, a natural baseline additionally considers remaining vehicle capacity when selecting customers, returning to the depot when no feasible customer can be served. The LLM-designed function is expected to discover more sophisticated selection strategies that go beyond local distance, potentially incorporating features such as angular sweep ordering, look-ahead insertion cost, demand-to-distance ratios, distance to the depot, or density-aware clustering of unvisited customers.





We apply the constructive framework to the two main-paper routing domains and,
for supplementary analysis, OVRP-Constructive. The detailed settings of the
LLM-designed selection function for each domain are as follows:

\begin{itemize}
    \item For \textbf{TSP-Constructive}, the LLM designs a function to select the next city to visit based on the current node, a destination node (node~0), the set of unvisited nodes, and the distance matrix. There is no capacity constraint and no depot return during construction. The function signature is $f(v,\;v_{\mathrm{dst}},\;\mathcal{U},\;D)\to v^\star$. All instances use 2D Euclidean coordinates sampled uniformly from $[0,1]^2$. The design set uses $N=50$ with 64 instances; validation uses $N\in\{50,100,200\}$ with 64 instances per size.
    \item For \textbf{CVRP-Constructive}, the LLM designs a function to select the next customer or decide to return to the depot, taking into account the current node, the depot location, feasible unvisited customers, remaining vehicle capacity, customer demands, and the distance matrix. The vehicle capacity is $Q=40$, customer demands are integer-valued and sampled uniformly from $\{1,\ldots,9\}$, and all coordinates (including the depot) are sampled uniformly from $[0,1]^2$. Instance sizes follow TSP-Constructive.
    \item For \textbf{OVRP-Constructive}, the function interface and instance generation are the same as CVRP-Constructive ($Q=40$, demands in $\{1,\ldots,9\}$, random depot). The only structural difference is that the final vehicle does not return to the depot, so the last return-edge cost is omitted from the objective. This domain is used only for supplementary held-out evaluation and is not counted among the eight main-paper domains.
\end{itemize}

\subsubsection{Ant Colony Optimization (ACO) Framework}

Ant Colony Optimization (ACO)~\citep{ACO:dorigo2006ant} is a meta-heuristic and population-based algorithm inspired by the foraging behavior of natural ant colonies. In ACO, a colony of $m$ artificial ants independently constructs candidate solutions in parallel. Each ant builds a solution step by step, selecting the next element with a probability that depends on two factors: (1)~a \emph{pheromone matrix} $\tau$, which encodes the collective search experience accumulated from previous iterations, and (2)~a \emph{heuristic desirability matrix} $\eta$, which provides problem-specific prior knowledge about the attractiveness of each transition.

The transition probability from node $i$ to node $j$ for ant $k$ is given by:
\[
p_{ij}^{(k)}=\frac{\tau_{ij}^\alpha\cdot\eta_{ij}^\beta\cdot\mathbb{1}[j\in\mathcal{N}_k]}
{\sum_{l\in\mathcal{N}_k}\tau_{il}^\alpha\cdot\eta_{il}^\beta},
\]
where $\mathcal{N}_k$ denotes the set of feasible candidates for ant $k$ (e.g., unvisited cities in TSP, or capacity-feasible customers in CVRP), and $\alpha,\beta$ control the relative influence of pheromone and heuristic information. The pheromone matrix is initialized uniformly ($\tau_{ij}=1$ for all $i,j$) and updated after each iteration: all values decay by a factor $\rho$ (evaporation), and edges appearing in high-quality solutions receive additional pheromone deposits proportional to the inverse solution cost. This feedback mechanism enables the colony to progressively concentrate its search around promising solution structures while maintaining exploratory diversity.

The key component designed by the LLM is the heuristic desirability matrix $\eta\in\mathbb{R}^{n\times n}$. Unlike the pheromone matrix, which evolves during optimization, the heuristic matrix is computed once from the instance data before the first iteration and remains fixed throughout. For example, the classical choice for TSP is $\eta_{ij}=1/D_{ij}$ (inverse distance)~\citep{TSPACO:skinderowicz2022improving}, which biases ants toward nearby cities but ignores global structure. The LLM can design more effective heuristic functions that incorporate features such as $k$-nearest-neighbor density, distance rank normalization, spatial clustering, or non-linear transformations that reshape the exploration-exploitation balance of the colony.

We apply the ACO framework to five main-paper domains, following the setting in
ReEvo~\cite{ReEvo:ye2024reevo} and the offline BPP ACO host used by
DeepACO~\cite{DeepACO:ye2023deepaco}. The LLM-designed heuristic function
$h$ is the sole component the model controls; its input signature varies by
domain, but the output is always a matrix (or vector for MKP) of non-negative
desirability values. The detailed settings for each domain are as follows:

For \textbf{TSP-ACO}, the heuristic function signature is $h(D)\to\eta$. We use $m=30$ ants, $T=100$ iterations, decay $\rho=0.9$, and exponents $\alpha=\beta=1$. Instances are 2D Euclidean with coordinates in $[0,1]^2$. The design set uses $N=50$ with 16 instances; validation uses $N\in\{50,100,200\}$ with 64 instances per size.

For \textbf{CVRP-ACO}, the heuristic function receives additional inputs: $h(D,X,d,Q)\to\eta$, where $X$ is the node coordinate matrix, $d$ is the demand vector, and $Q=50$ is the vehicle capacity. Customer demands are integer-valued and sampled uniformly from $\{1,\ldots,9\}$, and the depot is fixed at $(0.5,0.5)$. The ACO construction process incorporates capacity constraints: at each step, an ant can only transition to a customer whose demand does not exceed the remaining vehicle capacity, or return to the depot to start a new route. The remaining ACO parameters ($m=30$, $T=100$, $\rho=0.9$, $\alpha=\beta=1$) are the same as TSP-ACO. The design set uses $N=50$ with 10 instances; validation uses $N\in\{50,100,200\}$ with 64 instances per size.

For \textbf{OP-ACO}, the objective is to maximize the total prize collected within a maximum route length $L$. The heuristic function signature is $h(p,D,L)\to\eta$, where $p$ is the prize vector. Prizes are computed as $p_i = 1 + \lfloor 99 \cdot d_{0i}/\max_j d_{0j}\rfloor$ and normalized to $[0,1]$, where $d_{0i}$ is the distance from the depot. The maximum route length depends on problem size: $L=3.0$ for $N\le 50$, $L=4.0$ for $N\le 100$, $L=5.0$ for $N\le 200$, and $L=6.0$ for $N\le 300$. We use $m=20$ ants and $T=50$ iterations. This domain is used only for held-out evaluation; validation uses $N\in\{50,100,200\}$ with 64 instances per size.

For \textbf{MKP-ACO}, the Multidimensional Knapsack Problem has $m_{\mathrm{dim}}=5$ resource constraints. The heuristic function signature is $h(v,W)\to\eta$, where $v\in\mathbb{R}^n$ is the value vector and $W\in\mathbb{R}^{n\times 5}$ is the weight matrix. The ACO construction sequentially selects items, masking those that would violate any capacity constraint. Pheromone deposits are scaled by the total prize sum: $\Delta\tau = (1/\sum_i v_i)\cdot\text{objective}$. We use $m=10$ ants and $T=50$ iterations. This domain is used only for held-out evaluation; validation uses $N\in\{100,200,300\}$ with 5 instances per size.

For \textbf{BPP-ACO}, the heuristic function is
$h(s,C)\to\eta\in\mathbb{R}^{n\times n}$, where $s$ is the item-size
vector and $\eta_{ij}$ is the desirability of grouping items $i$ and $j$ in
the same bin. Each ant repeatedly selects a capacity-feasible unpacked item for
the current bin and opens a new bin when no remaining item fits. We use 20 ants,
pheromone decay $0.95$, and $\alpha=\beta=1$. Each design-time heuristic
evaluation runs 15 ACO iterations on 5 $N=500$ instances; final validation runs
100 ACO iterations on 64 instances at each of
$N\in\{500,1000,2000\}$. Item sizes are uniform integers in
$\{20,\ldots,100\}$ and $C=150$. This domain is excluded from RL training.

\subsubsection{Bayesian Optimization (BO) Framework}

Bayesian optimization (BO)~\citep{BO:shahriari2015taking} is a method for optimizing expensive black-box functions where the objective is to find the global minimum of an unknown function $f(x)$ over a bounded search space $\mathcal{X}$:
\[
x^\star = \arg\min_{x\in\mathcal{X}} f(x).
\]
Two core components of BO are the \emph{probabilistic surrogate model} and the \emph{acquisition function}~\citep{movckus1974bayesian}. In each iteration, the surrogate model (typically a Gaussian process) is fitted to all observed evaluations, providing a posterior distribution with mean $\hat{\mu}(x)$ and standard deviation $\hat{\sigma}(x)$ at any candidate point $x$. The acquisition function then uses these posterior statistics to determine the next evaluation point, balancing exploitation (evaluating where the predicted objective is good) and exploration (evaluating where uncertainty is high).

In cost-aware BO settings, different evaluation points incur different costs, and the total evaluation budget is measured in cumulative cost rather than number of evaluations. This requires the acquisition function to additionally consider the predicted cost $\hat{c}(x)$ of each candidate, managing the trade-off between expected improvement, uncertainty, and evaluation cost under a fixed budget. The goal of LLM-based AHD in this setting is to design a \emph{cost-aware acquisition function} (CAF) that can adaptively balance these three objectives. The LLM designs a utility function $u$ that replaces the standard acquisition function and receives the full surrogate state---posterior mean, standard deviation, cost prediction, training history, and remaining budget---and returns a scalar value that, when maximized, guides the search toward promising and cost-efficient regions.

Standard acquisition functions such as Expected Improvement (EI)~\cite{Bo-CAF:yao2024evolve} use a closed-form formula that depends only on $\hat{\mu}(x)$, $\hat{\sigma}(x)$, and the current best value. Cost-aware variants like EI-per-unit-cost~\citep{EIpu:snoek2012practical} divide EI by the predicted cost raised to a decaying power: $a(x)=\mathrm{EI}(x)/\hat{c}(x)^\alpha$, where $\alpha$ decreases as the remaining budget shrinks. These formulas encode fixed trade-off strategies that cannot adapt to the specific geometry of the objective landscape. The LLM-designed utility function is expected to discover acquisition strategies that go beyond these templates, potentially conditioning on the full training history, adapting the exploration-exploitation ratio based on observed progress, or incorporating cost-awareness in non-linear ways.

For \textbf{CAF}, the cost-aware BO benchmark uses 12 standard test functions with varying input dimensions: Ackley ($d=2$), Rastrigin ($d=2$), Griewank ($d=2$), Rosenbrock ($d=2$), Levy ($d=2$), ThreeHumpCamel ($d=2$), StyblinskiTang ($d=2$), Hartmann-3D ($d=3$), Powell ($d=2$), Shekel ($d=4$), Hartmann-6D ($d=6$), and Cosine8 ($d=8$). The evaluation cost function is $c(x)=\exp(-\|x-x_{\mathrm{opt}}\|_2)$, which is highest near the optimum and decreases with distance from it; this makes evaluations near the global optimum expensive, incentivizing the acquisition function to balance proximity to the optimum against evaluation cost. The total cost budget is $C_{\max}=30$.


\section{Definition of AHD Agent}
\label{appendix:definition-ahd-agent}

This section gives the implementation-level definition of \methodname{} used in
our experiments. We separate the agent-facing diagnostic tools from the
evaluation components required to execute candidate heuristics. Diagnostic tools
provide design-set-only information to the model during the design loop, whereas the
evaluation components manage sessions, execute submitted code on design
instances, and record the resulting feedback.

\subsection{Diagnostic Tool Interfaces}
\label{appendix:tool-details}

All diagnostic tools are design-set-only: they may inspect the current design
instances, the session-local attempt history, and evaluator feedback already
produced during the design loop, but they never expose validation or test
objectives. We focus on two diagnostic interfaces used by the agent:
\texttt{InstanceAnalysis} and \texttt{ASTNoveltyAnalyzer}.

\begin{table*}[t]
\centering
\footnotesize
\setlength{\tabcolsep}{4pt}
\renewcommand{\arraystretch}{1.16}
\caption{\textbf{Diagnostic tool interface summary.} The diagnostic tools are
read-only, operate only on the design set and session history, and do not
consume evaluator calls.}
\label{tab:tool-interface-summary}
\begin{tabular}{@{}p{0.21\textwidth}p{0.25\textwidth}p{0.30\textwidth}p{0.18\textwidth}@{}}
\toprule
Tool & Reads & Returns & Design-loop role \\
\midrule
\texttt{InstanceAnalysis} &
\texttt{session\_id}; \texttt{scope} =
\texttt{summary}, \texttt{single\_instance}, or
\texttt{contrastive\_pair}; optional \texttt{instance\_id}. &
Dataset-level or instance-level feature summaries, including spatial,
demand, density, and contrastive statistics when available. &
Helps the agent form dataset hypotheses before spending evaluator calls. \\
\addlinespace[2pt]
\texttt{ASTNoveltyAnalyzer} &
\texttt{session\_id}; candidate \texttt{code}; optional \texttt{top\_k}. &
AST similarity to previous attempts, novelty band,
nearest historical matches, and an evaluation hint. &
Flags near-duplicates before evaluation. \\
\bottomrule
\end{tabular}
\end{table*}

\subsubsection{Design-Instance Analysis}

\textbf{InstanceAnalysis.}
This tool summarizes the design set bound to the current task. It requires a \texttt{session\_id} and supports three scopes. With \texttt{scope=summary}, it analyzes the full design set and returns aggregated feature statistics. With \texttt{scope=single\_instance}, it additionally requires an \texttt{instance\_id} and returns a detailed feature summary for that one design instance. With \texttt{scope=contrastive\_pair}, it selects the most dissimilar pair of design instances by standardized feature distance and reports the largest feature gaps between them. The output is a text summary for the model and a structured metrics dictionary for logging.

\textbf{Feature computation.}
For each instance in the design set, the tool computes five categories of structural features from the raw coordinates and domain-specific attributes:
\begin{itemize}
    \item \textbf{Spacing uniformity}: using a KD-tree, the tool computes each node's nearest-neighbor distance. It reports the coefficient of variation of these distances (\texttt{nn\_cv}) and the ratio of the observed mean nearest-neighbor distance to the expected value under a uniform random distribution (\texttt{nn\_mean\_normalized}). These metrics indicate whether nodes are regularly spaced, clustered, or randomly scattered.
    \item \textbf{Cluster structure}: DBSCAN is applied with $\varepsilon$ set to the 10th percentile of nearest-neighbor distances. The tool reports the number of detected clusters and the silhouette score when at least two clusters are found. This helps the agent identify whether the instance has distinct spatial groups that may benefit from cluster-aware heuristic logic.
    \item \textbf{Density variation}: a 2D histogram grid partitions the coordinate space, and the coefficient of variation of bin counts (\texttt{density\_cv}) quantifies how unevenly nodes are distributed across the spatial domain.
    \item \textbf{Boundary shape}: the convex hull of the node coordinates is computed. The tool reports the fraction of nodes on the hull (\texttt{hull\_fraction}) and the ratio of hull area to bounding-box area (\texttt{hull\_area\_ratio}), characterizing whether nodes fill the interior or concentrate near the boundary.
    \item \textbf{Demand pattern} (CVRP/OVRP only): for domains with per-customer demands, the tool computes the demand coefficient of variation (\texttt{demand\_cv}) and Moran's~I spatial autocorrelation index (\texttt{demand\_morans\_i}) using 5-nearest-neighbor spatial weights. Moran's~I reveals whether high-demand customers are spatially clustered or randomly distributed, which can inform capacity-aware routing strategies.
\end{itemize}

\textbf{Aggregation.}
When \texttt{scope=summary}, the tool computes the above features for every instance in the design set and aggregates them by reporting the mean, minimum, and maximum of each metric. The aggregated statistics are then translated into a natural-language summary returned to the agent. This summary allows the agent to form hypotheses about the dataset structure---such as whether instances are clustered, whether demands are spatially correlated, or whether boundary nodes are disproportionately represented---before spending evaluator calls.

\subsubsection{Code-Structure Analysis}

\textbf{ASTNoveltyAnalyzer.}
The AST novelty tool compares a candidate heuristic program against all previously evaluated attempts in the same session. Its inputs are \texttt{session\_id}, the candidate \texttt{code}, and an optional \texttt{top\_k} (default 3). The tool does not decide which heuristic is best; it only exposes structural similarity signals so that the agent can judge whether another evaluator call is worthwhile or whether further revision is needed first.

\textbf{AST normalization.}
Comparing raw source code is unreliable because superficial differences---such as variable renaming, constant tuning, or comment changes---inflate dissimilarity without reflecting genuine algorithmic changes. To address this, the tool uses Python's \texttt{ast} module to parse both the candidate and each historical attempt into abstract syntax trees. It then applies a structure-preserving normalization pass that replaces all variable names with a placeholder \texttt{VAR}, all function arguments with \texttt{ARG}, and all literal constants with type-level placeholders (\texttt{NUM} for numbers, \texttt{STR} for strings, \texttt{BOOL} for booleans). The resulting normalized tree---referred to as the \emph{shape tree}---captures only the control-flow and operator structure of the program, discarding surface-level variation.

\textbf{Similarity computation.}
The tool computes three complementary similarity scores between the candidate and each historical attempt:
\begin{itemize}
    \item \textbf{Raw similarity}: the \texttt{SequenceMatcher} ratio between the full (unnormalized) AST dumps of the two programs. This component is sensitive to variable names and constant values, capturing cases where the candidate is a near-verbatim copy.
    \item \textbf{Shape similarity}: the \texttt{SequenceMatcher} ratio between the normalized shape-tree dumps. This component ignores naming and constant differences and focuses on whether the control-flow structure (branches, loops, function calls) has changed.
    \item \textbf{Node-type similarity}: the cosine similarity between the node-type frequency vectors of the two ASTs (e.g., counts of \texttt{If}, \texttt{For}, \texttt{Call}, \texttt{BinOp}, etc.). This component captures coarse structural composition even when the tree layout differs.
\end{itemize}
The final AST similarity is a weighted combination:
\[
s = 0.25 \times s_{\mathrm{raw}} + 0.50 \times s_{\mathrm{shape}} + 0.25 \times s_{\mathrm{node}}.
\]
The shape component receives the highest weight because structural changes to control flow and operator composition are the strongest indicators of genuinely different algorithmic logic.

The tool returns the AST similarity to each of the top-$k$ most similar historical attempts, along with a candidate summary (node count, branches, loops, constants, function calls), a novelty score ($1 - s_{\max}$), and an evaluation hint that helps the agent decide whether to evaluate or revise further.

\subsubsection{Diagnostic Tool-Use Constraints}

The diagnostic tools are intentionally indirect. They expose instance
structure and code-structure novelty, but they do not rank candidates by
validation performance. The agent must decide whether these signals justify a
new design-time evaluation or a code revision. This design prevents validation
leakage while still allowing the policy to perform model-controlled
experimentation, diagnose failure modes, and revise heuristics over multiple
turns.

\subsection{Evaluation Implementation}
\label{appendix:evaluation-implementation}

Candidate execution is implemented by the evaluation harness rather than by the
diagnostic tool set. Each design run starts by creating a persistent session
workspace identified by a \texttt{session\_id}. The workspace stores the
problem configuration, the design-set binding, all candidate programs
submitted during the run, execution logs, aggregate objective values, and
instance-level cost histories. This persistent state makes the interaction
history available to later diagnostic calls without exposing validation data.

When the agent submits a candidate heuristic, the harness writes the complete
Python implementation into an isolated session-local evaluation directory. The
same evaluator used by the corresponding baseline solver then executes the
candidate on the design instances. The harness records whether the program is
executable and feasible, its average design-set objective, repeat-level statistics
when repeated evaluation is enabled, a monotonically increasing attempt id, and
the best-so-far design-set objective in the current session. For successful
evaluations, the harness also persists per-instance costs by problem size. These
records support session auditing and debugging without exposing validation data.

The evaluator budget is counted only when a candidate is actually executed on
the design set. Diagnostic calls are read-only and do not consume evaluator
calls. During search, the agent observes only design-time feedback: execution
errors, feasibility status, design-set objective values, best-so-far status,
diagnostic summaries, and the remaining evaluator budget. Validation and test
sets are used only after the final heuristic is selected, so the design loop
does not receive validation or test feedback.

\subsection{Prompt Templates}
\label{appendix:prompts}

This section summarizes the prompt templates used by the multi-turn heuristic
designer. Unlike fixed-search frameworks that maintain separate prompts for
initialization, mutation, crossover, or tree-path reasoning actions,
\textbf{\methodname{}} uses one system prompt and one user prompt. The task
identity, function interface, seed code, and observations are injected through
placeholders such as \texttt{\{problem.description\}},
\texttt{\{function\_signature\}}, \texttt{\{initial\_code\}}, and
\texttt{\{algorithm\_details(given\_heuristics)\}}.

\subsubsection{System Prompt}

The system prompt defines the role of the model, the diagnostic tools, and the
global interaction protocol. The same template is shared by all problem domains;
only \texttt{\{task\_brief\}}, \texttt{\{objective\_text\}}, and the available
diagnostic interfaces are instantiated at runtime.

\begin{tcolorbox}[
  colback=red!2!white,
  colframe=red!55!black,
  title={System prompt template},
  fonttitle=\bfseries,
  breakable,
  left=4pt,right=4pt,top=4pt,bottom=4pt]
\small
\textbf{Role.} You are an expert in designing heuristic algorithms for
\texttt{\{task\_brief\}}. Your goal is to iteratively improve the current
heuristic and optimize \texttt{\{objective\_text\}}.

\textbf{Available diagnostic interfaces.}
\begin{enumerate}
  \item \texttt{InstanceAnalysis}: summarize structural properties of the
  design instances, such as spacing, clustering, density, boundary statistics,
  and task-specific attributes when available.
  \item \texttt{ASTNoveltyAnalyzer}: compare the AST structure of a candidate
  against previously evaluated candidates. This interface is used only as a
  novelty checkpoint; final ranking is always determined by design-time evaluation.
\end{enumerate}

\textbf{Interaction rules.} Use diagnostic feedback and design-time evaluation
results to revise the code over multiple turns. Do not submit the initial code
unchanged. The final answer must contain only the complete Python code after
the marker \texttt{\#\#\#\# FINAL SOLUTION \#\#\#\#}.
\end{tcolorbox}

\subsubsection{User Prompt Template}

The user prompt contains the task-specific design request, the current
baseline, the implementation constraints, appended observations, and the
final-answer format. For constructive routing tasks, the target is a next-node
decision function; for ACO tasks, the target is a heuristic-information
function; for CAF, the target is an acquisition utility function. These
differences are represented by placeholders rather than by separate prompt
families.

\begin{tcolorbox}[
  colback=blue!2!white,
  colframe=blue!55!black,
  title={User prompt template},
  fonttitle=\bfseries,
  breakable,
  left=4pt,right=4pt,top=4pt,bottom=4pt]
\small
\textbf{Task request.} Please design the required heuristic function using the
following background.

\textbf{Problem description.} \texttt{\{problem.description\}}

\textbf{Algorithmic context.} \path|{algorithm_details(given_heuristics)}|

\textbf{Function interface.} \texttt{\{function\_signature\}}

\textbf{Current baseline code.} \texttt{\{initial\_code\}}

\textbf{Current baseline objective.} \texttt{\{baseline\_objective\}}

\textbf{Evaluation objective.} Direction: \texttt{\{objective\_direction\}}.
Improve over the current baseline using only design-set feedback.

\textbf{Implementation constraints.} The function name and signature must match
\texttt{\{function\_name\}} exactly. The returned value must satisfy all
validity constraints in \texttt{\{problem.description\}}. The code must be
deterministic and executable in the provided evaluator. NumPy is available as
\texttt{np}.

\textbf{Observations appended during interaction.} After each diagnostic call
or design-time evaluation, the returned observation is appended to the conversation
memory. The agent may then revise a candidate, call a diagnostic interface,
submit a new implementation for design-time evaluation, or finalize the best
successfully evaluated code.

\textbf{Final answer format.} All domains use the same final-answer constraint:
\texttt{\#\#\#\# FINAL SOLUTION \#\#\#\# \textless complete Python code only\textgreater}.
\end{tcolorbox}

\subsubsection{Problem Information Used in Prompts}

Table~\ref{tab:prompt-problem-info} summarizes the domain-specific information
inserted into \texttt{\{problem.description\}} and related placeholders. The
entries are intentionally concise because the contribution of this work is the
agentic interaction policy rather than domain-specific prompt engineering.
The first eight rows are the main-paper domains. OVRP-Constructive is listed
last and retained only as an additional supplementary evaluation domain.

\begin{table*}[t]
\centering
\small
\setlength{\tabcolsep}{4pt}
\caption{Information of each problem used in prompts. The same system and user
prompt templates are reused across domains; only these problem-specific fields
are instantiated.}
\label{tab:prompt-problem-info}
\resizebox{\textwidth}{!}{
\begin{tabular}{llll}
\toprule
Problem & Target function & Unit / direction & Prompt description \\
\midrule
TSP-Constructive & Next-node selector & Tour length ($\downarrow$) & Construct a tour that visits each node exactly once and returns to the start node. The heuristic chooses the next node during step-by-step route construction. \\
CVRP-Constructive & Next-node selector & Travel distance ($\downarrow$) & Serve customers with known demands using capacitated vehicles while minimizing total route length and respecting vehicle capacity. \\
TSP-ACO & Heuristic matrix & Tour length ($\downarrow$) & Design heuristic information for ACO so that shorter and structurally promising TSP edges receive higher transition desirability. \\
CVRP-ACO & Heuristic matrix & Travel distance ($\downarrow$) & Design ACO heuristic information for customer transitions under vehicle-capacity constraints, balancing short travel and route feasibility. \\
OP-ACO & Heuristic matrix & Collected reward ($\uparrow$) & Select a subset of reward-bearing locations under a travel budget. The heuristic estimates transition desirability for reward collection within the budget. \\
MKP-ACO & Heuristic score & Packed profit ($\uparrow$) & Select items for a multidimensional knapsack under multiple capacity constraints. The heuristic estimates item desirability for ACO-style solution construction. \\
BPP-ACO & Item-pair heuristic matrix & Number of bins ($\downarrow$) & Pack a known set of one-dimensional items into the minimum number of fixed-capacity bins. The heuristic estimates whether pairs of items should be grouped in the same bin during ACO construction. \\
CAF & Acquisition utility & Optimization gap ($\downarrow$) & Design a cost-aware Bayesian-optimization acquisition utility that balances predicted improvement, uncertainty, and evaluation cost under a fixed BO budget. \\
OVRP-Constructive (supplementary) & Next-node selector & Travel distance ($\downarrow$) & Construct open vehicle routes that serve all customers without requiring each route to return to the depot, while minimizing total travel distance. \\
\bottomrule
\end{tabular}
}
\end{table*}

\section{RL Training Details}
\label{appendix:rl-training}

This section provides additional details on the RL training process, including
training data statistics, training configuration, and training curves.

\subsection{Training Data}
\label{appendix:rl-training-data}

The \textit{AHD Environment Synthesis Pipeline} subsection in the main paper
describes the training data synthesis pipeline at a conceptual level. Here we
report the concrete statistics of the final training set.

\begin{figure*}[t]
\centering
\includegraphics[width=0.92\textwidth]{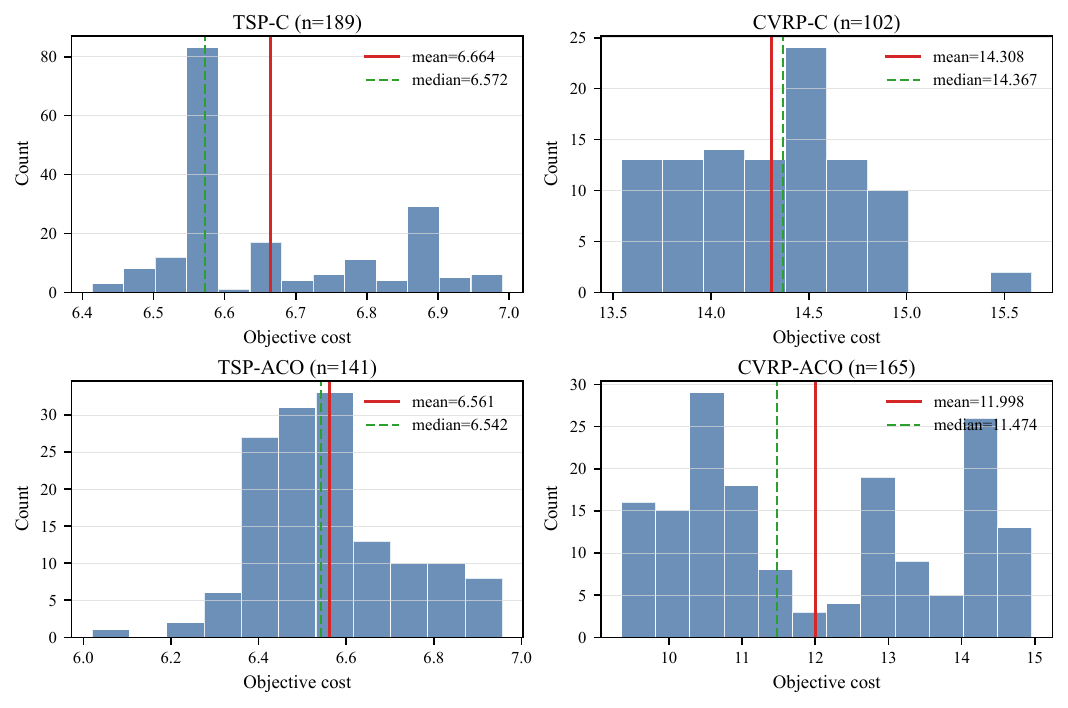}
\caption{Objective-value distribution of the source heuristic pools used to generate the RL training data. Solid and dashed vertical lines mark the pool mean and median, respectively. All four domains exhibit broad coverage from near-baseline to competitive quality levels.}
\label{fig:rl-training-cost-distribution}
\end{figure*}

The RL training data is constructed from seed heuristics collected via ReEvo searches and prior agentic design sessions across four domains. For each domain, we retain all executable heuristics from the collection process---not only the best-performing ones---to ensure diversity in starting quality. Each seed heuristic is paired with multiple training-instance variants at problem sizes $N\in\{40,50,60,70,80\}$, and the baseline objective is re-evaluated on each variant. Failed or timed-out evaluations are discarded. The final dataset contains 1{,}000 candidate prompts per domain, drawn from the source pools summarized in Table~\ref{tab:rl-training-data-summary}. Approximately 50\% of the prompts use the \emph{improve-from-code} mode, and the remaining 50\% use the \emph{design-from-scratch} mode. Figure~\ref{fig:rl-training-cost-distribution} visualizes the objective-value distribution of the source heuristic pools, showing that all four domains cover a broad range of starting quality levels.

\begin{table*}[t]
\centering
\small
\setlength{\tabcolsep}{4pt}
\caption{Training data statistics for the four-domain RL dataset. Source heuristics are the number of distinct seed programs collected per domain. Generated prompts are candidate RL records after pairing with instance variants. Training sizes report the instance scales used for variant augmentation, and the format ratio reports improve-from-code versus design-from-scratch prompts. Obj.\ value statistics are computed over the source pool.}
\label{tab:rl-training-data-summary}
\resizebox{\textwidth}{!}{
\begin{tabular}{lrrccrr}
\toprule
Domain & Source heuristics & Generated prompts & Train sizes $N$ & Improve:Scratch & Mean obj.\ value & Median obj.\ value \\
\midrule
TSP-Constructive & 189 & 1{,}000 & $\{40,50,60,70,80\}$ & 1:1 & 6.664 & 6.572 \\
CVRP-Constructive & 102 & 1{,}000 & $\{40,50,60,70,80\}$ & 1:1 & 14.308 & 14.367 \\
TSP-ACO & 141 & 1{,}000 & $\{40,50,60,70,80\}$ & 1:1 & 6.561 & 6.542 \\
CVRP-ACO & 165 & 1{,}000 & $\{40,50,60,70,80\}$ & 1:1 & 11.998 & 11.474 \\
\midrule
Total & 597 & 4{,}000 & $\{40,50,60,70,80\}$ & 1:1 & -- & -- \\
\bottomrule
\end{tabular}
}
\end{table*}

\subsection{Training Configuration}
\label{appendix:rl-training-setup}

The RL policy is initialized from Qwen3-4B-Instruct-2507~\citep{Qwen3:qwen3technicalreport} and trained with GRPO in the multi-turn tool-augmented environment described. Each RL step samples a mini-batch of 4 prompts, generates 8 rollouts per prompt, executes tool calls and evaluations within the environment, and updates the policy using the staged reward described in the \textit{Reward Design} subsection of the main paper. Training runs for 500 steps on 4$\times$A100-80G GPUs. Table~\ref{tab:rl-training-config} summarizes the full hyperparameter configuration.

\begin{table}[t]
\centering
\small
\setlength{\tabcolsep}{5pt}
\caption{RL training hyperparameters.}
\label{tab:rl-training-config}
\resizebox{\columnwidth}{!}{
\begin{tabular}{ll}
\toprule
Hyperparameter & Value \\
\midrule
Base model & Qwen3-4B-Instruct-2507~\citep{Qwen3:qwen3technicalreport} \\
Training algorithm & GRPO~\citep{shao2024deepseekmath} \\
Training domains & TSP-C, CVRP-C, TSP-ACO, CVRP-ACO \\
Prompts per step (batch size) & 4 \\
Rollouts per prompt & 8 \\
Learning rate & $1\times10^{-6}$ \\
KL coefficient $\beta$ & 0.001 \\
Max prompt length & 5{,}120 tokens \\
Max response length & 44{,}032 tokens \\
Validation frequency & every 50 steps \\
Total training steps & 500 \\
Hardware & 4$\times$A100-80G, ${\sim}$140 GPU-hours \\
\bottomrule
\end{tabular}
}
\end{table}

\subsection{Training Curves}
\label{appendix:rl-training-curves}

\begin{figure}[t]
\centering
\begin{minipage}{0.48\columnwidth}
\centering
\includegraphics[width=\linewidth]{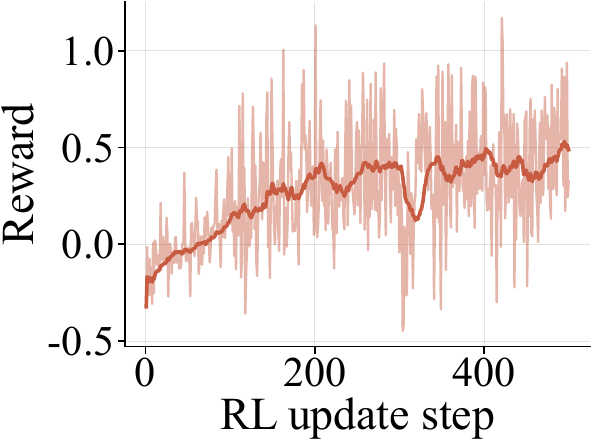}
\end{minipage}\hfill
\begin{minipage}{0.48\columnwidth}
\centering
\includegraphics[width=\linewidth]{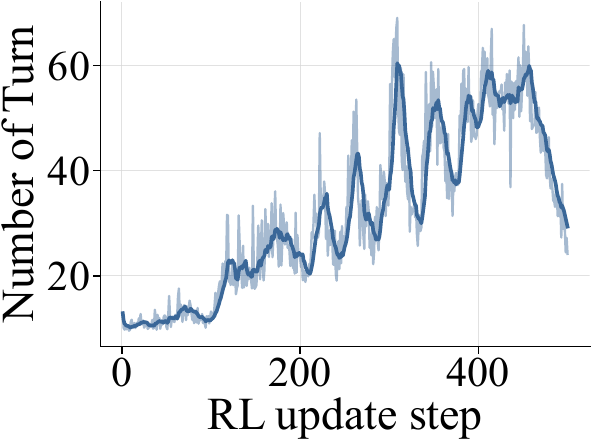}
\end{minipage}

\vspace{0.6em}

\begin{minipage}{0.48\columnwidth}
\centering
\includegraphics[width=\linewidth]{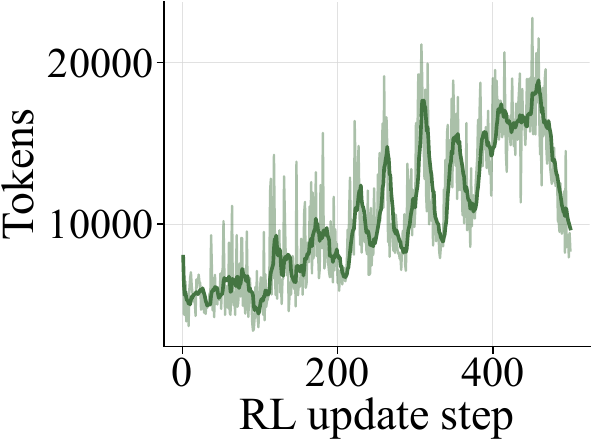}
\end{minipage}\hfill
\begin{minipage}{0.48\columnwidth}
\centering
\includegraphics[width=\linewidth]{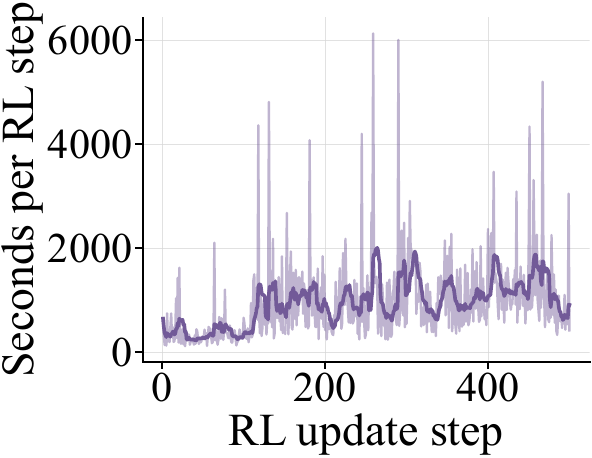}
\end{minipage}
\caption{RL training diagnostics over 500 steps. Top-left: quality reward (higher is better). Top-right: average interaction turns per rollout. Bottom-left: average response length in tokens. Bottom-right: wall-clock time per RL step.}
\label{fig:rl-training-diagnostics}
\end{figure}

\begin{figure}[t]
\centering
\begin{minipage}{0.48\columnwidth}
\centering
\includegraphics[width=\linewidth]{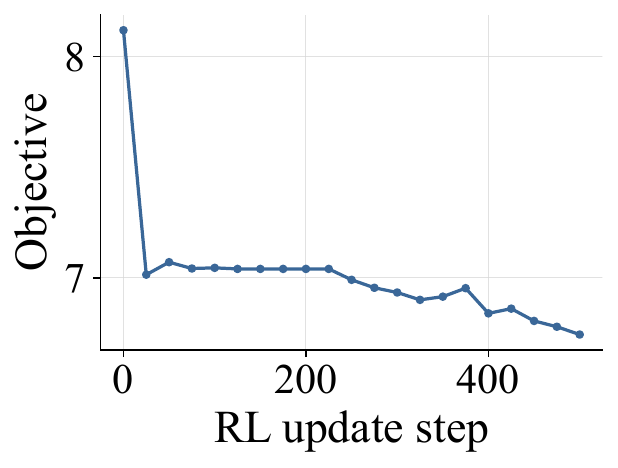}
\end{minipage}\hfill
\begin{minipage}{0.48\columnwidth}
\centering
\includegraphics[width=\linewidth]{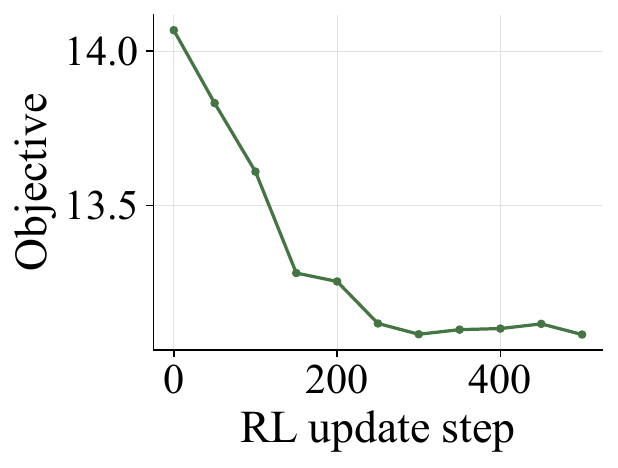}
\end{minipage}

\vspace{0.6em}

\begin{minipage}{0.48\columnwidth}
\centering
\includegraphics[width=\linewidth]{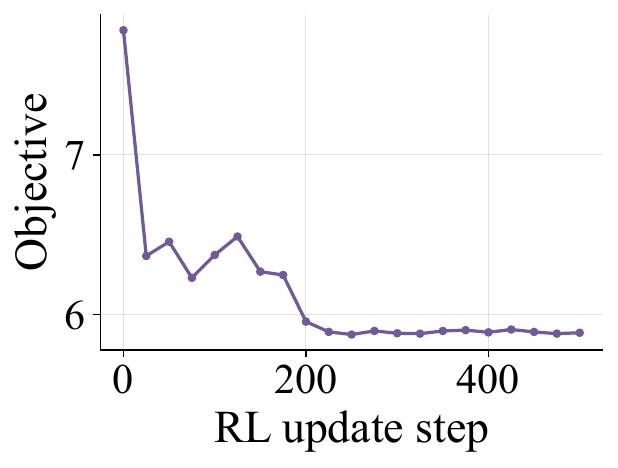}
\end{minipage}\hfill
\begin{minipage}{0.48\columnwidth}
\centering
\includegraphics[width=\linewidth]{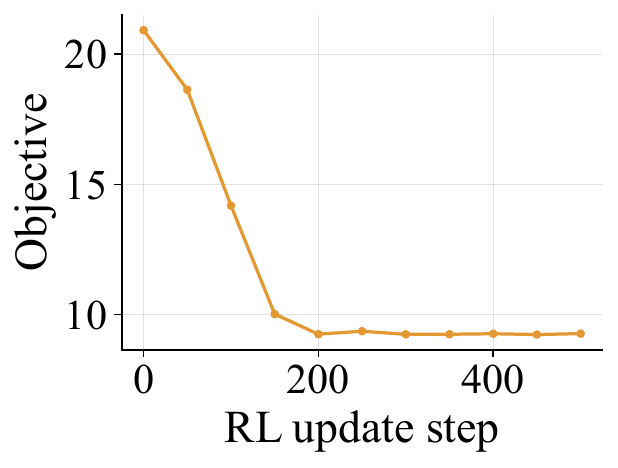}
\end{minipage}
\caption{Per-domain validation curves over RL training steps. Each panel reports the $N=50$ validation objective; lower is better for all four domains.}
\label{fig:rl-training-validation}
\end{figure}

Figures~\ref{fig:rl-training-diagnostics} and \ref{fig:rl-training-validation} report key training diagnostics and per-domain validation curves over 500 RL steps.

The quality reward (Figure~\ref{fig:rl-training-diagnostics}, top-left) increases steadily throughout training, indicating that the policy progressively learns to produce higher-quality heuristics. The average number of interaction turns (top-right) stabilizes after an initial adjustment period, suggesting that the policy converges to a consistent multi-turn interaction strategy rather than expanding trajectory length indefinitely. Response length (bottom-left) shows a moderate increase as the model learns to generate more detailed reasoning and code revisions. Step time (bottom-right) remains stable, confirming that the training infrastructure scales consistently across steps.

The per-domain validation curves (Figure~\ref{fig:rl-training-validation}) show that the validation objective improves on all four domains over the course of training. TSP-ACO and CVRP-ACO converge relatively quickly, while CVRP-Constructive shows more gradual improvement, consistent with the higher variance of constructive heuristic design.

\section{Detailed Experimental Results}
\label{appendix:detailed-results}

\subsection{Pricing Details}
\label{appendix:pricing}

API-based backbones use the corresponding OpenRouter rates. Locally served
Qwen3-4B and RL checkpoints are assigned the Qwen3-8B proxy rate used by the
local pricing entry.

\begin{table*}[t]
\centering
\small
\setlength{\tabcolsep}{5pt}
\caption{Token pricing used for the reported cost estimates. Prices are in USD
per one million tokens.}
\label{tab:appendix-token-pricing}
\resizebox{0.8\textwidth}{!}{
\begin{tabular}{lcc}
\toprule
Model used in tables & Input price & Output price \\
\midrule
DeepSeek-V4-Flash & \$0.14/M & \$0.28/M \\
GPT-4o & \$2.50/M & \$10.00/M \\
Qwen3-8B proxy for local Qwen/RL models & \$0.05/M & \$0.40/M \\
\bottomrule
\end{tabular}
}
\end{table*}

\subsection{Metrics Details}
\label{app:reference_seed}

This section provides the details behind the rows labeled ``Optimal'' and ``Baseline heuristic'' in Tables~\ref{tab:table20-rl-training-mean-gap-no-para} and~\ref{tab:table21-generalization-mean-gap-no-para}. 
The ``Optimal'' row reports the reference objective $f^\star$ used to compute validation Gap, while the ``Baseline heuristic'' row reports the performance of the initial heuristic provided to each heuristic-design method. 
For all LLM-based AHD methods compared within the same domain, we use the same target function interface, design instances, validation instances, objective computation, and initial baseline heuristic when applicable; only the search or interaction workflow differs.

\paragraph{Gap computation.}
We use validation Gap (\%) as the primary performance metric. 
For minimization domains, including TSP-Constructive, CVRP-Constructive,
TSP-ACO, CVRP-ACO, BPP-ACO, CAF, and the supplementary OVRP-Constructive
domain, the Gap is computed as
\[
\mathrm{Gap} = \frac{f - f^\star}{|f^\star|} \times 100\%,
\]
where $f$ is the objective value achieved by the evaluated heuristic and $f^\star$ is the corresponding optimal or best-known reference objective. 
For maximization domains, including OP-ACO and MKP-ACO, the Gap is computed as
\[
\mathrm{Gap} = \frac{f^\star - f}{|f^\star|} \times 100\%.
\]
Under this convention, lower Gap is always better, and a Gap of $0$ indicates that the evaluated heuristic matches the reference objective on the validation set.

\textbf{Reference objectives.}
For each validation domain and problem size, the reference objective $f^\star$ is computed independently of the AHD search process and is never exposed to the agent during design. 
For TSP-Constructive and TSP-ACO, the reference tour lengths are computed by LKH~\cite{LKH:helsgaun2017extension}. 
For CVRP-Constructive, CVRP-ACO, and OVRP-Constructive, the reference routing objectives are computed by PyVRP~\cite{PYVRP:wouda2024pyvrp}. 
For OP-ACO, the reference collected rewards are computed by OP-Solver~\cite{OP-Solver:kobeaga2024revisited}. 
For MKP-ACO, the reference packed profits are computed by OR-Tools~\cite{ortools}. 
For BPP-ACO, we compute an exact optimum for every validation instance. The
capacity lower bound $\lceil\sum_i s_i/C\rceil$ is matched by a certified
feasible packing on every instance, which proves optimality; the ``Optimal''
row reports the corresponding per-size mean exact optimum.
These reference objectives are used only for reporting validation Gap after a final heuristic has been selected.

\textbf{Baseline heuristics.}
The initial baseline heuristic defines the starting point of the improve-from-code setting and also provides the ``Baseline heuristic'' row in the main tables. 
For TSP-Constructive, the baseline follows the classical greedy constructive rule~\cite{TSPGC:rosenkrantz1977analysis}. 
For CVRP-Constructive, we use the same constructive starting point as LLM4AD~\cite{LLM4AD:liu2024llm4ad}. 
For OVRP-Constructive, we use the same constructive rule as CVRP-Constructive, adapted to the open-route objective by omitting the final return-to-depot cost. 
For TSP-ACO, CVRP-ACO, OP-ACO, and MKP-ACO, the baseline heuristic follows the standard ReEvo ACO settings~\cite{ReEvo:ye2024reevo}. 
For BPP-ACO, the initial heuristic is
$\eta_{ij}=s_j/\max_k s_k$, replicated across rows; the ``Baseline
heuristic'' row reports standard ACO with this heuristic over five random seeds
using 20 ants and 100 iterations, following the offline BPP ACO
host~\cite{DeepACO:ye2023deepaco}.
In ACO domains, the LLM-designed component is the heuristic desirability function, while the remaining ACO solver configuration is kept fixed across methods.

\textbf{Evaluation protocol.}
All reported method results are computed after the design process terminates and the final heuristic is selected. 
During design, the agent and all baselines receive feedback only from the design instances; validation and test objectives are not available to the search process. 
The evaluator-call count reports the number of times candidate heuristic code is executed on the design set. 
Diagnostic tool calls do not consume evaluator budget, and Cost reports the average USD cost per run under the pricing protocol described in Section~\ref{appendix:pricing}. 
Unless otherwise specified, method results, evaluator calls, and costs are averaged over five independent runs.

\subsection{Data Sources and Split Protocol}
\label{app:data_split_protocol}

Our experiments involve three different data sources, which serve different purposes and are never used interchangeably: the RL prompt corpus, the evaluation-time design set, and the held-out validation set.

\paragraph{RL prompt corpus.}
The RL prompt corpus $\mathcal{Q}_{\mathrm{RL}}$ is used only to optimize the parameters of the AHD Agent policy with GRPO. Each RL prompt instantiates a train-time design environment, including a problem domain, a seed heuristic or a design-from-scratch mode, a design-dataset variant, the tool library, and the reward/evaluation interface. These prompts are used to train the agentic interaction policy, i.e., how the model uses feedback, calls tools, revises code, repairs invalid candidates, and decides when to finalize. The RL prompt corpus is not used for reporting the validation results in the main tables.

\paragraph{Evaluation-time design set.}
During evaluation, every AHD run is associated with a design set $D_{\mathrm{design}}$. This set is the only set visible to the heuristic-design loop. Candidate heuristics generated by AHD Agent or by fixed-workflow AHD baselines are evaluated on $D_{\mathrm{design}}$ during search, and all design-time evaluator scores, execution feedback, and per-instance diagnostic records are computed only on this set. Diagnostic tools are also restricted to $D_{\mathrm{design}}$, the current session history, and evaluator feedback already produced within the same design run. They do not inspect held-out validation instances, validation objectives, or validation per-instance performance. The design set is used to select the final heuristic within the allowed evaluator-call budget, but design-set performance is not the metric reported in the main result tables.

\paragraph{Held-out validation set.}
After the design budget is exhausted, the selected heuristic is fixed and evaluated on the held-out validation set $D_{\mathrm{val}}$. No validation feedback is returned to the agent, tools, evaluator loop, or fixed-workflow search process during heuristic design. Therefore, the reported COP results measure held-out validation performance of a heuristic selected using only design-set feedback. For the combinatorial optimization domains in this paper, $D_{\mathrm{val}}$ is the final reporting split; we do not use an additional COP test split. The cost-aware Bayesian optimization benchmark follows its own test-function protocol described in Appendix~\ref{appendix:problem-domains}.

\paragraph{In-domain and held-out domains.}
The distinction between in-domain and held-out domains refers only to whether a domain appears in the RL prompt corpus used to train the AHD Agent policy. In-domain domains are included in $\mathcal{Q}_{\mathrm{RL}}$, whereas held-out domains are absent from RL training. However, both in-domain and held-out evaluation domains still use an evaluation-time design set $D_{\mathrm{design}}$ during AHD search, because all AHD methods require design-time feedback to discover a heuristic. Thus, held-out-domain experiments test whether the trained agentic design policy transfers to new problem families, not whether the final heuristic is produced without any design-time feedback.

\begin{table*}[t]
\centering
\caption{Instance split statistics for COP domains. ``Used in RL training'' indicates whether the domain appears in the GRPO prompt corpus. Held-out domains are not used to train the AHD Agent policy, but they still have an evaluation-time design set for heuristic search. BPP-ACO belongs to the seven-domain main-paper combinatorial benchmark; OVRP-Constructive is an additional supplementary domain.}
\label{tab:instance_split_statistics}
\resizebox{\textwidth}{!}{%
\begin{tabular}{lllll}
\toprule
\textbf{Domain} & \textbf{Used in RL training?} & \textbf{Evaluation-time design set} & \textbf{Held-out validation set} & \textbf{Instance distribution} \\
\midrule
TSP-Constructive & Yes & $N=50$, 64 instances & $N\in\{50,100,200\}$, 64 instances per size & 2D Euclidean, coordinates sampled uniformly from $[0,1]^2$ \\
CVRP-Constructive & Yes & $N=50$, 64 instances & $N\in\{50,100,200\}$, 64 instances per size & 2D Euclidean depot/customer coordinates sampled uniformly from $[0,1]^2$; integer demands in $\{1,\ldots,9\}$; capacity $Q=40$ \\
TSP-ACO & Yes & $N=50$, 16 instances & $N\in\{50,100,200\}$, 64 instances per size & 2D Euclidean, coordinates sampled uniformly from $[0,1]^2$ \\
CVRP-ACO & Yes & $N=50$, 10 instances & $N\in\{50,100,200\}$, 64 instances per size & 2D Euclidean CVRP distribution; depot fixed at $(0.5,0.5)$; integer demands in $\{1,\ldots,9\}$; capacity $Q=50$ \\
OP-ACO & No & $N=50$, 5 instances & $N\in\{50,100,200\}$, 64 instances per size & 2D Euclidean OP distribution; prizes are distance-derived; route-length budget depends on instance size \\
MKP-ACO & No & $N=100$, 5 instances & $N\in\{100,200,300\}$, 5 instances per size & Multidimensional knapsack with $m=5$ resource constraints; prizes and weights sampled uniformly and capacities normalized \\
BPP-ACO & No & $N=500$, 5 instances & $N\in\{500,1000,2000\}$, 64 instances per size & Offline 1D bin packing; item sizes sampled uniformly from integer $\{20,\ldots,100\}$; bin capacity $C=150$ \\
OVRP-Constructive (supplementary) & No & $N=50$, 64 instances & $N\in\{50,100,200\}$, 64 instances per size & CVRP-style distribution with open routes; integer demands in $\{1,\ldots,9\}$; capacity $Q=40$ \\
\bottomrule
\end{tabular}
}
\end{table*}

\subsection{Motivation for RL Training}

\textbf{General-purpose LLMs are backbone-sensitive.}
Table~\ref{tab:rl-motivation-backbone-instability} compares GPT-4o and
DeepSeek-V4-Flash under the same fixed-search AHD framework on an auxiliary
seven-domain subset: six main-paper combinatorial domains and the additional
OVRP-Constructive domain. BPP-ACO is not included in this legacy backbone
sweep. The best backbone varies by domain and framework: for example, on
TSP-ACO EOH performs better with GPT-4o, while on CVRP-ACO it performs better
with DeepSeek-V4-Flash. ReEvo shows a similar inconsistency, with the dominant
backbone flipping between domains. This instability means that practitioners
must re-evaluate backbone choices for each new domain, adding cost and
uncertainty to the design process.

\begin{table*}[t]
\centering
\tiny
\setlength{\tabcolsep}{3.0pt}
\caption{Model-sensitivity view for motivating RL-trained multi-turn
design. Entries are average validation mean Gap (\%) over the reported sizes,
comparing GPT-4o (G) with DeepSeek-V4-Flash (D) under the same AHD method. Each
cell reports the lower-gap model first, with the other model in parentheses;
lower is better. This auxiliary subset contains six main-paper combinatorial
domains plus the supplementary OVRP-Constructive domain; BPP-ACO was not part
of this legacy backbone sweep.}
\label{tab:rl-motivation-backbone-instability}
\resizebox{\textwidth}{!}{
\begin{tabular}{lccccccc}
\toprule
Method & TSP-C & CVRP-C & TSP-ACO & CVRP-ACO & OP-ACO & OVRP-C & MKP-ACO \\
\midrule
ReEvo
& \textbf{D} 17.64 (G 23.95)
& \textbf{D} 25.60 (G 29.31)
& \textbf{D} 8.15 (G 11.94)
& \textbf{D} 35.16 (G 168.03)
& \textbf{G} 11.71 (D 12.44)
& \textbf{D} 106.12 (G 111.26)
& \textbf{D} 2.04 (G 4.20) \\
EOH
& \textbf{G} 17.68 (D 22.78)
& \textbf{D} 26.59 (G 31.06)
& \textbf{G} 8.54 (D 9.38)
& \textbf{G} 23.81 (D 26.59)
& \textbf{D} 9.85 (G 11.04)
& \textbf{D} 107.85 (G 108.72)
& \textbf{D} 1.78 (G 2.76) \\
MCTS-AHD
& \textbf{G} 19.45 (D 20.68)
& $\approx$27.0 / 27.0
& \textbf{G} 9.11 (D 9.46)
& \textbf{D} 27.98 (G 38.45)
& \textbf{D} 9.24 (G 10.50)
& \textbf{D} 106.89 (G 107.03)
& \textbf{G} 2.56 (D 2.63) \\
\bottomrule
\end{tabular}
}
\end{table*}

\textbf{RL-trained \methodname{} is efficient with a fixed compact local backbone.}
Table~\ref{tab:motivation-rl-efficiency} compares \methodname{} (RL) against
fixed-search AHD baselines with GPT-4o on four representative domains: two
training domains, the main-paper held-out OP-ACO domain, and the additional
supplementary OVRP-Constructive domain. \methodname{} achieves the lowest
average gap (32.85\%) across all four domains while using only 12.7 evaluator
calls on average, compared to 100 calls for the fixed-search baselines. The
average cost per run is \$0.004, roughly two orders of magnitude lower than the
GPT-4o baselines (\$0.31--\$0.85). At inference time, the fixed Qwen3-4B
checkpoint avoids repeated per-domain selection among costly frontier API
backbones.

\begin{table*}[t]
\centering
\scriptsize
\setlength{\tabcolsep}{2.0pt}
\caption{Efficiency summary on four representative domains (two in-domain,
one main-paper held-out domain, and the supplementary OVRP-Constructive domain).
For each domain, Gap~(\%) is the average Mean~Gap over the reported validation sizes
($N=100,200$ for training domains; $N=50,100,200$ for held-out domains), matching
the main-text tables. Eval is the average evaluator-call count per run. Cost is the
average USD per run. Lower is better for all metrics.}
\label{tab:motivation-rl-efficiency}
\resizebox{\textwidth}{!}{
\begin{tabular}{lrrrrrrrrrrrrrrr}
\toprule
 & \multicolumn{6}{c}{Training Domains} & \multicolumn{6}{c}{Held-out Domains} & \multicolumn{3}{c}{Overall} \\
\cmidrule(lr){2-7} \cmidrule(lr){8-13} \cmidrule(lr){14-16}
 & \multicolumn{3}{c}{CVRP-C} & \multicolumn{3}{c}{TSP-ACO} & \multicolumn{3}{c}{OP-ACO} & \multicolumn{3}{c}{OVRP-C} & \multicolumn{3}{c}{Average} \\
\cmidrule(lr){2-4} \cmidrule(lr){5-7} \cmidrule(lr){8-10} \cmidrule(lr){11-13} \cmidrule(lr){14-16}
Method & Gap & Eval & Cost & Gap & Eval & Cost & Gap & Eval & Cost & Gap & Eval & Cost & Gap & Eval & Cost \\
\midrule
\methodname{} (RL) & \textbf{19.77} & \textbf{14.6} & \textbf{0.004} & \textbf{9.95} & \textbf{11.8} & \textbf{0.003} & \textbf{8.93} & \textbf{13.6} & \textbf{0.005} & \textbf{92.75} & \textbf{10.7} & \textbf{0.004} & \textbf{32.85} & \textbf{12.7} & \textbf{0.004} \\
EOH + GPT-4o & 29.64 & 100 & 0.332 & 11.22 & 100 & 0.236 & 11.04 & 100 & 0.343 & 108.72 & 100 & 0.346 & 40.15 & 100.0 & 0.314 \\
MCTS-AHD + GPT-4o & 25.58 & 100 & 0.702 & 11.91 & 100 & 1.026 & 10.50 & 100 & 0.930 & 107.03 & 100 & 0.728 & 38.75 & 100.0 & 0.847 \\
ReEvo + GPT-4o & 27.69 & 100 & 0.642 & 15.65 & 100 & 0.473 & 11.71 & 100 & 0.903 & 111.26 & 100 & 0.678 & 41.58 & 100.0 & 0.674 \\
\bottomrule
\end{tabular}
}
\end{table*}

\subsection{Comparison with PathWise}
\label{appendix:pathwise-comparison}

We additionally evaluate the recently introduced PathWise
framework~\citep{PathWise:gungordu2026pathwise} on all seven combinatorial
domains using DeepSeek-V4-Flash. PathWise and the fixed-search baselines use
100 design-time evaluator calls per run, and each result is averaged over five
independent runs. To match the main-paper reporting protocol, Table
~\ref{tab:pathwise-seven-domain-comparison} averages Gap over $N=100,200$ for
the four training domains and over the three sizes reported in Table~2 for
OP-ACO, MKP-ACO, and BPP-ACO.

\begin{table*}[t]
\centering
\small
\setlength{\tabcolsep}{3.0pt}
\caption{Seven-domain comparison with PathWise. Entries are average validation
Gap (\%) over the problem sizes reported in main-paper Tables~1 and~2; lower is
better. Eval is the average number of design-time evaluator calls per run.
Results are averaged over five independent runs.}
\label{tab:pathwise-seven-domain-comparison}
\resizebox{\textwidth}{!}{%
\begin{tabular}{lccccccccc}
\toprule
Method & TSP-C & CVRP-C & TSP-ACO & CVRP-ACO & OP-ACO & MKP-ACO & BPP-ACO & Average & Eval \\
\midrule
\multicolumn{10}{c}{\textbf{LLM-based AHD: DeepSeek-V4-Flash}} \\
\midrule
EOH & 23.50 & 26.38 & 12.32 & 29.39 & 9.85 & 1.78 & 1.96 & 15.02 & 100.0 \\
MCTS-AHD & 21.73 & 26.34 & 12.45 & 30.09 & 9.24 & 2.63 & 1.85 & 14.90 & 100.0 \\
ReEvo & 18.89 & 24.83 & 10.89 & 39.12 & 12.44 & 2.04 & 2.20 & 15.77 & 100.0 \\
\methodname{} & 22.45 & 22.96 & 10.60 & 45.60 & 12.56 & 1.81 & 2.99 & 16.99 & 18.8 \\
PathWise & 16.93 & 26.45 & 9.93 & 29.98 & 14.94 & 1.79 & 2.41 & 14.63 & 100.0 \\
\midrule
\multicolumn{10}{c}{\textbf{RL Model: Qwen3-4B-Instruct-2507}} \\
\midrule
CALM & 18.83 & 35.45 & 14.60 & 33.01 & 10.60 & 2.49 & 4.70 & 17.09 & 150.0 \\
\rowcolor{gray!15}\methodname{} & 18.02 & 19.77 & 9.96 & 33.61 & 8.93 & 1.40 & 2.77 & 13.49 & 12.8 \\
\rowcolor{gray!15}\methodname{} w/SR
& \textbf{14.50} & \textbf{19.68} & \textbf{9.80} & \textbf{24.40}
& \textbf{7.08} & \textbf{0.90} & \textbf{1.85} & \textbf{11.17} & 100.0 \\
\bottomrule
\end{tabular}
}
\end{table*}

Under the same 100-evaluation budget, \methodname{} w/SR outperforms PathWise
on every domain. Its seven-domain average Gap is 11.17\%, compared with
14.63\% for PathWise, an improvement of 3.46 percentage points. The short
\methodname{} setting also achieves a lower overall Gap (13.49\%) while using
only 12.8 evaluator calls on average. These results show that the learned
agentic design policy remains more effective than the newly introduced
fixed-workflow method under both matched-budget and efficiency-oriented
comparisons.

\subsection{Model Scaling with GPT-Series Models}
\label{appendix:gpt-scaling}

\begin{figure*}[t]
\centering
\includegraphics[width=0.92\textwidth]{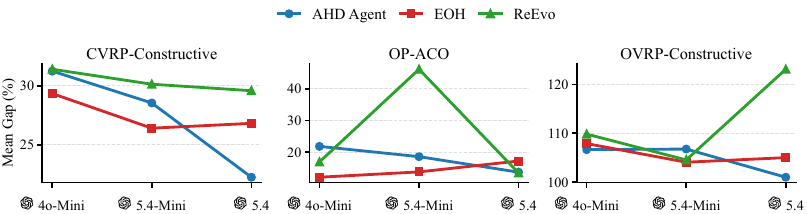}
\caption{Mean validation Gap (\%) across the three reported problem sizes as
the GPT-series backbone changes from GPT-4o-Mini to GPT-5.4-Mini and GPT-5.4. \methodname{} uses the LLM as an agent that iteratively calls
tools and reacts to feedback, whereas EOH and ReEvo use the LLM inside a fixed
search procedure.}
\label{fig:backbone-gap-scaling}
\end{figure*}

In addition to the Qwen-family scaling, we conduct the same comparison using
three GPT-series models (GPT-4o-Mini, GPT-5.4-Mini, GPT-5.4) on
CVRP-Constructive, OP-ACO, and OVRP-Constructive. OVRP-Constructive is the
additional supplementary domain outside the eight-domain main-paper benchmark.
Figure~\ref{fig:backbone-gap-scaling} reports the results.


\textbf{\methodname{} shows a clear scaling trend.} The agentic multi-turn framework consistently improves with stronger backbones: on all three domains, the best gap is achieved with the strongest model GPT-5.4, with endpoint improvements of $-9.00$, $-8.14$, and $-5.65$ percentage points from GPT-4o-Mini to GPT-5.4. This monotonic trend confirms that the multi-turn paradigm can effectively convert stronger model capabilities into better heuristic design decisions.

\textbf{Fixed-search frameworks do not exhibit a consistent scaling trend.} EOH and ReEvo show non-monotonic or domain-dependent behavior under the same backbone sweep. On OP-ACO, EOH degrades from 12.06\% at GPT-4o-Mini to 17.18\% at GPT-5.4. ReEvo is even more erratic: its OP-ACO gap spikes to 46.16\% at GPT-5.4-Mini before recovering, and its OVRP-Constructive gap increases by $+13.26$ points when moving to the strongest backbone. These results suggest that simply using a more powerful LLM as a code generator inside a fixed search procedure does not reliably improve performance.

\subsection{Tool Ablation with DeepSeek-V4-Flash}

Tools consistently benefit the agentic multi-turn framework but produce mixed or negative effects in fixed-search workflows (Table~\ref{tab:deepseek-v4-tool-ablation-two-domain}). This experiment uses DeepSeek-V4-Flash as the LLM backbone for all three frameworks, comparing two tool configurations: \textbf{A0} (evaluator only---the agent can evaluate candidates but has no diagnostic tools) and \textbf{A2} (full tools---instance analysis and AST novelty are additionally available). We report mean and best validation gaps averaged over three problem sizes ($N=50,100,200$) on CVRP-Constructive and TSP-ACO.

\textbf{\methodname{} results.} Adding diagnostic tools (A0$\to$A2) consistently improves the multi-turn agent on both domains. The mean gap decreases by 1.87 points on CVRP-Constructive (25.60$\to$23.73) and by 1.15 points on TSP-ACO (9.30$\to$8.15). The best gap also improves: $-1.73$ points on CVRP-Constructive and $-0.10$ points on TSP-ACO. This confirms that the agentic multi-turn policy can effectively leverage additional diagnostic signals to guide its revisions.

\textbf{Fixed-workflow tool injection.} To provide the same diagnostic information to EOH and ReEvo, we inject the tools through deterministic adapters that preserve the original evolutionary workflow:
\begin{itemize}
    \item \textbf{EOH}: A drop-in \emph{ToolAugmentedSampler} wraps the original EoH sampler with a four-phase pipeline. (1)~Instance analysis computes a structural summary of the design instances and appends it to the code-generation prompt. (2)~The base sampler generates a candidate as usual. (3)~AST novelty analysis compares the candidate against the current population. (4)~If the candidate is structurally too similar to existing individuals, the LLM is asked to revise it once before evaluation.
    \item \textbf{ReEvo}: A \emph{ReEvoToolController} injects the instance summary (computed once and cached) into the LLM prompts before initialization, crossover, and mutation operations. After each candidate is generated, the controller performs AST novelty checking against the evolutionary history and triggers a single revision if similarity exceeds a threshold.
\end{itemize}
In both cases, tool injection is \emph{fixed and deterministic}: the LLM cannot choose whether or when to query the tools, and the feedback is always appended in the same format at the same pipeline stage.

\textbf{Fixed-workflow results.} Unlike the multi-turn agent, EOH and ReEvo show inconsistent responses to tool injection. On CVRP-Constructive, EOH improves only marginally (mean gap $-0.27$) and ReEvo improves moderately ($-1.19$). On TSP-ACO, both frameworks \emph{degrade} with tools: EOH worsens by $+0.18$ and ReEvo by $+0.49$ on mean gap. This asymmetry demonstrates that the same diagnostic information can be helpful or harmful depending on how it is integrated. When tools are injected at fixed points without the model's agency over when and how to use the feedback, the additional information may disrupt the existing search dynamics rather than improve them. The multi-turn agent, by contrast, can choose to query tools selectively, ignore uninformative feedback, and adapt its revision strategy based on the diagnostic results.

\begin{table*}[t]
\centering
\scriptsize
\setlength{\tabcolsep}{3.2pt}
\newcommand{\improvecell}[1]{\colorbox{gray!15}{\strut #1}}
\caption{DeepSeek-V4-Flash tool ablation on two domains. Entries are mean/best
validation Gap (\%) averaged over three sizes; lower is better. \methodname{}
compares evaluator-only tools with AST/analysis tools, while EOH and ReEvo
compare the original design with AST/analysis tools. Shading marks improved
full-tool cells; bold and underline mark the best and second-best values.}
\label{tab:deepseek-v4-tool-ablation-two-domain}
\resizebox{0.72\textwidth}{!}{
\begin{tabular}{cccccc}
\toprule
\multicolumn{1}{c}{\multirow{2}{*}{Framework}} & \multicolumn{1}{c}{\multirow{2}{*}{Setting}} & \multicolumn{2}{c}{CVRP-C} & \multicolumn{2}{c}{TSP-ACO} \\
\cmidrule(lr){3-4} \cmidrule(lr){5-6}
& & Mean Gap (\%) & Best Gap (\%) & Mean Gap (\%) & Best Gap (\%) \\
\midrule
\multirow{3}{*}{\methodname{}} & Only Eval Tool & 25.597 & \underline{22.423} & 9.298 & 7.628 \\
& With AST/Analysis Tools & \improvecell{\textbf{23.728}} & \improvecell{\textbf{20.691}} & \improvecell{\underline{8.146}} & \improvecell{7.524} \\
& Improvement & 1.869 & 1.732 & 1.152 & 0.104 \\
\midrule
\multirow{3}{*}{EOH} & Ori-Design & 26.586 & 25.637 & 9.384 & 7.216 \\
& With AST/Analysis & \improvecell{26.317} & 25.637 & 9.562 & 7.770 \\
& Improvement & 0.269 & 0.000 & -0.178 & -0.554 \\
\midrule
\multirow{3}{*}{ReEvo} & Ori-Design & 25.603 & 23.176 & \underline{8.154} & \textbf{7.150} \\
& With AST/Analysis & \improvecell{\underline{24.411}} & \improvecell{22.701} & 8.643 & 7.956 \\
& Improvement & 1.192 & 0.475 & -0.489 & -0.806 \\
\bottomrule
\end{tabular}
}
\end{table*}

\subsection{\methodname{} w/PS vs.\ \methodname{} w/SR: Budget-Expanded Inference}

The standard \methodname{} setting uses a short evaluator budget (up to 30 calls) to maximize efficiency. When a larger evaluator budget is available, we study two complementary strategies for scaling multi-turn inference: \textbf{Sequential Refinement} (\methodname{} w/SR) and \textbf{Parallel Sampling} (\methodname{} w/PS). Both strategies reuse the same RL-trained policy without any additional fine-tuning; they differ only in how the evaluator budget is allocated across interaction trajectories.

\subsubsection{Sequential Refinement (\methodname{} w/SR)}

\methodname{} w/SR increases inference depth by extending a single agent session across multiple continuation rounds until a global evaluator budget (default 100 calls) is exhausted. Concretely, the agent begins with its standard multi-turn interaction. When it submits a \texttt{FINAL SOLUTION}\ marker but the global budget has not been reached, the system automatically starts a new continuation round: the best-evaluated heuristic from previous rounds is injected as the new seed code, a continuation note informs the agent of the remaining budget, and the conversation history is reset to avoid context-length overflow. This process repeats for up to $K$ dialog rounds (default $K=10$) or until the global budget is fully consumed.

The key advantage of this strategy is that the agent accumulates design experience across rounds. Each continuation round starts from the best result of the previous round, enabling progressive refinement of a single heuristic trajectory. The agent can exploit insights from earlier evaluations---such as which code patterns improved performance or which instance groups remain weak---to guide subsequent revisions.

\subsubsection{Parallel Sampling (\methodname{} w/PS)}

\methodname{} w/PS increases inference breadth by launching \(L\)\ independent short multi-turn lanes in parallel (default $L=5$). Each lane runs a complete standard \methodname{} session with its own conversation context and evaluator state. The lanes execute concurrently using a thread pool and do not share information during execution. After all lanes complete, the system selects the best candidate across all lanes based on the design-set objective.

This strategy trades depth for diversity. While each individual lane uses only a short budget, the parallel ensemble samples $L$ independent trajectories from the policy, increasing the probability that at least one trajectory discovers a strong heuristic. The total evaluator budget is approximately $L$ times the per-lane budget. \methodname{} w/PS is naturally parallelizable and more robust to individual trajectory failures: if one lane produces a weak or failed candidate, the remaining lanes can still succeed.

\subsubsection{Comparison}

Table~\ref{tab:pipe-vs-long-multiturn-summary} compares both strategies on
the seven domains for which both variants are available: six main-paper
combinatorial domains plus the supplementary OVRP-Constructive domain. BPP-ACO
is excluded from this comparison because a corresponding \methodname{} w/PS
result is not reported. \methodname{} w/SR obtains the lower average mean gap
on five of these seven domains (TSP-C, TSP-ACO, CVRP-ACO, OP-ACO, MKP-ACO),
while \methodname{} w/PS is better on two (CVRP-C, OVRP-C). Averaged across
this explicitly defined subset, \methodname{} w/SR achieves a mean gap of
23.57\%, compared with 24.47\% for \methodname{} w/PS. \methodname{} w/PS
uses 88.8 evaluator calls on average, compared with the fixed 100-call budget
for \methodname{} w/SR. In practice, the choice between the two strategies
depends on the deployment scenario: \methodname{} w/SR is preferred when
serial depth and progressive refinement are important, while \methodname{}
w/PS is preferred when wall-clock time is constrained and parallelism is
available.

\begin{table}[t]
\centering
\small
\setlength{\tabcolsep}{3.0pt}
\caption{\methodname{} w/PS versus \methodname{} w/SR on the seven
domains where both budget-expanded variants are reported: six main-paper
combinatorial domains plus the supplementary OVRP-Constructive domain. BPP-ACO
is omitted because no corresponding w/PS result is reported. \methodname{}
w/PS corresponds to the \methodname{} w/PS rows in the appendix mean-gap
tables. For each domain, Avg.\ Mean Gap averages the three validation sizes.
Avg.\ Eval reports the design-time evaluator budget: \methodname{} w/PS uses
the reported evaluator-call count averaged over the same sizes, while
\methodname{} w/SR uses a fixed 100-call sequential-refinement budget. Lower
is better.}
\label{tab:pipe-vs-long-multiturn-summary}
\resizebox{\columnwidth}{!}{%
\begin{tabular}{lcccc}
\toprule
Domain & \multicolumn{2}{c}{Avg.\ Mean Gap (\%)} & \multicolumn{2}{c}{Avg.\ Eval} \\
\cmidrule(lr){2-3} \cmidrule(lr){4-5}
 & PS & SR & PS & SR \\
\midrule
TSP-Constructive & 14.80 & \textbf{13.62} & \textbf{100.0} & \textbf{100.0} \\
CVRP-Constructive & \textbf{18.85} & 21.05 & 107.6 & \textbf{100.0} \\
TSP-ACO & 7.65 & \textbf{7.57} & \textbf{99.6} & 100.0 \\
CVRP-ACO & 26.32 & \textbf{22.01} & 126.3 & \textbf{100.0} \\
OP-ACO & 8.45 & \textbf{7.08} & \textbf{38.6} & 100.0 \\
OVRP-Constructive & \textbf{92.70} & 92.75 & 122.6 & \textbf{100.0} \\
MKP-ACO & 2.53 & \textbf{0.90} & \textbf{26.6} & 100.0 \\
\midrule
Average & 24.47 & \textbf{23.57} & \textbf{88.8} & 100.0 \\
\bottomrule
\end{tabular}
}
\end{table}

\subsection{P-values for Significance}
\label{appendix:pvalue-significance}

Following the significance-test protocol in prior AHD work~\citep{MCTS-AHD:zheng2025monte}, we conduct ten independent runs for each method and test whether the RL-trained multi-turn agent significantly outperforms the strongest LLM-based AHD baseline on each domain. Each run is summarized by the average validation mean gap over three problem sizes ($N\in\{50,100,200\}$). We compare the run-level gap distributions using single-tailed Welch $t$-tests, where the alternative hypothesis is that the multi-turn agent has a lower mean gap than the baseline. For each domain, we select the best-performing AHD baseline with DeepSeek-V4-Flash as the comparison target.

Table~\ref{tab:pvalue-significance-ahd} reports the per-run gaps, averages, standard deviations, and $p$-values. On \textbf{CVRP-Constructive}, the standard RL-trained \methodname{} (avg.\ 22.13\%) significantly outperforms the best baseline ReEvo (avg.\ 25.44\%) with $p=4.65\times 10^{-5}$. On \textbf{OVRP-Constructive}, an additional supplementary held-out domain outside the eight-domain main-paper benchmark, the RL-trained \methodname{} (avg.\ 92.75\%) likewise significantly outperforms ReEvo (avg.\ 106.30\%) with $p=1.44\times 10^{-9}$, demonstrating transfer to this auxiliary domain.

On \textbf{OP-ACO}, the standard RL-trained \methodname{} (avg.\ 8.93\%) is slightly better than the best baseline MCTS-AHD (avg.\ 9.22\%) on average, but the difference is not statistically significant ($p=0.355$) due to higher run-to-run variance under the limited short multi-turn evaluator budget. When we increase the evaluator budget with Sequential Refinement \methodname{} (\methodname{} w/SR), the variance shrinks substantially (std from 1.73 to 0.52) and the gap drops to 7.08\%, achieving significance with $p=7.80\times 10^{-4}$. This confirms that the RL-trained policy has learned an effective design strategy; the additional evaluator budget allows it to realize this advantage more consistently.

\begin{table*}[t]
\centering
\caption{Ten-run significance tests against strong AHD baselines. Each run reports the mean gap
(\%) averaged over $N=50,100,200$. ``avg'' and ``std'' are computed over the ten run-level gaps,
and $p$-values are from single-tailed Welch $t$-tests. Lower is better.}
\label{tab:pvalue-significance-ahd}
\scriptsize
\setlength{\tabcolsep}{2.4pt}
\resizebox{\textwidth}{!}{%
\begin{tabular}{llrrrrrrrrrrrrr}
\toprule
Domain & Method & run1 & run2 & run3 & run4 & run5 & run6 & run7 & run8 & run9 & run10 & avg & std & $p$-value \\
\midrule
CVRP-C & ReEvo & 23.377 & 24.812 & 27.330 & 26.993 & 25.504 & 25.559 & 27.151 & 25.716 & 22.125 & 25.875 & 25.444 & 1.576 & -- \\
 & \methodname{} RL & 20.926 & 22.229 & 23.637 & 22.098 & 22.576 & 20.926 & 23.452 & 20.926 & 23.637 & 20.926 & 22.133 & 1.109 & $\mathbf{4.65{\times}10^{-5}}$ \\
OVRP-C & ReEvo & 102.632 & 105.358 & 110.180 & 104.906 & 107.519 & 107.022 & 105.167 & 106.235 & 108.805 & 105.161 & 106.299 & 2.054 & -- \\
 & \methodname{} RL & 92.549 & 92.549 & 94.227 & 92.549 & 92.549 & 92.549 & 92.865 & 92.549 & 92.549 & 92.549 & 92.749 & 0.502 & $\mathbf{1.44{\times}10^{-9}}$ \\
\midrule
OP-ACO & MCTS-AHD & 11.342 & 8.524 & 7.518 & 10.339 & 8.259 & 11.678 & 8.652 & 7.465 & 10.112 & 8.284 & 9.217 & 1.455 & -- \\
 & \methodname{} RL & 6.896 & 10.528 & 9.214 & 7.866 & 7.554 & 10.139 & 12.662 & 7.583 & 9.534 & 7.360 & 8.934 & 1.727 & $3.55{\times}10^{-1}$ \\
 & \methodname{} w/SR RL & 7.202 & 7.867 & 7.626 & 6.796 & 7.152 & 7.829 & 6.384 & 6.398 & 6.729 & 6.853 & 7.084 & 0.520 & $\mathbf{7.80{\times}10^{-4}}$ \\
\bottomrule
\end{tabular}
}
\end{table*}

\subsection{Detailed Seven-Domain Results with PathWise}
\label{appendix:pathwise-detailed-results}

Tables~\ref{tab:pathwise-detailed-in-domain}
and~\ref{tab:pathwise-detailed-held-out} expand the summary in
Table~\ref{tab:pathwise-seven-domain-comparison} into the per-size objective
Cost and Gap values used to compute each domain average. The problem sizes and
evaluation protocol match main-paper Tables~1 and~2.

\begin{table*}[t]
\centering
\scriptsize
\setlength{\tabcolsep}{1.2pt}
\caption{Detailed in-domain comparison with PathWise. Each problem size reports
the mean objective Cost and mean Gap (\%) over five independent runs. Eval is
the average number of design-time evaluator calls per run. Lower Gap is better.}
\label{tab:pathwise-detailed-in-domain}
\resizebox{\textwidth}{!}{%
\begin{tabular}{l*{20}{c}}
\toprule
& \multicolumn{5}{c}{TSP-Constructive ($\downarrow$)}
& \multicolumn{5}{c}{CVRP-Constructive ($\downarrow$)}
& \multicolumn{5}{c}{TSP-ACO ($\downarrow$)}
& \multicolumn{5}{c}{CVRP-ACO ($\downarrow$)} \\
\cmidrule(lr){2-6} \cmidrule(lr){7-11} \cmidrule(lr){12-16} \cmidrule(lr){17-21}
Method
& \multicolumn{2}{c}{N=100} & \multicolumn{2}{c}{N=200} & Eval
& \multicolumn{2}{c}{N=100} & \multicolumn{2}{c}{N=200} & Eval
& \multicolumn{2}{c}{N=100} & \multicolumn{2}{c}{N=200} & Eval
& \multicolumn{2}{c}{N=100} & \multicolumn{2}{c}{N=200} & Eval \\
\cmidrule(lr){2-3} \cmidrule(lr){4-5}
\cmidrule(lr){7-8} \cmidrule(lr){9-10}
\cmidrule(lr){12-13} \cmidrule(lr){14-15}
\cmidrule(lr){17-18} \cmidrule(lr){19-20}
& Cost & Gap & Cost & Gap &
& Cost & Gap & Cost & Gap &
& Cost & Gap & Cost & Gap &
& Cost & Gap & Cost & Gap & \\
\midrule
\multicolumn{21}{c}{\textbf{LLM-based AHD: DeepSeek-V4-Flash}} \\
\midrule
EOH
& 9.48 & 22.76\% & 13.29 & 24.23\% & 100
& 22.67 & 28.01\% & 40.22 & 24.75\% & 100
& 8.47 & 8.84\% & 12.37 & 15.80\% & 100
& 16.28 & 28.11\% & 29.20 & 30.66\% & 100 \\
MCTS-AHD
& 9.35 & 20.94\% & 13.11 & 22.52\% & 100
& 22.68 & 28.07\% & 40.14 & 24.61\% & 100
& 8.48 & 8.86\% & 12.40 & 16.04\% & 100
& 16.46 & 29.51\% & 29.20 & 30.66\% & 100 \\
ReEvo
& 9.09 & 17.70\% & 12.84 & 20.07\% & 100
& 22.43 & 26.61\% & 39.66 & 23.05\% & 100
& 8.35 & 7.26\% & 12.23 & 14.51\% & 100
& 17.31 & 36.23\% & 31.74 & 42.00\% & 100 \\
\methodname{}
& 9.45 & 22.29\% & 13.11 & 22.60\% & 18.4
& 22.08 & 24.60\% & 39.12 & 21.32\% & 20.4
& 8.39 & 7.77\% & 12.12 & 13.42\% & 14.2
& 18.33 & 44.25\% & 32.84 & 46.95\% & 17.6 \\
PathWise
& 8.94 & 15.78\% & 12.63 & 18.09\% & 100
& 22.44 & 26.74\% & 40.67 & 26.16\% & 100
& 8.33 & 7.02\% & 12.05 & 12.85\% & 100
& 16.20 & 27.46\% & 29.61 & 32.50\% & 100 \\
\midrule
\multicolumn{21}{c}{\textbf{RL Model: Qwen3-4B-Instruct-2507}} \\
\midrule
CALM
& 8.85 & 14.60\% & 13.16 & 23.05\% & 150
& 24.13 & 36.30\% & 43.24 & 34.59\% & 150
& 8.61 & 10.56\% & 12.67 & 18.63\% & 150
& 16.61 & 30.72\% & 30.24 & 35.30\% & 150 \\
\rowcolor{gray!15}\methodname{}
& 9.04 & 17.07\% & 12.72 & 18.96\% & 12.8
& 21.47 & 21.17\% & 38.17 & 18.37\% & 14.6
& 8.35 & 7.20\% & 12.04 & 12.71\% & 11.8
& 16.60 & 30.67\% & 30.52 & 36.54\% & 12.9 \\
\rowcolor{gray!15}\methodname{} w/SR
& 8.80 & 13.86\% & 12.31 & 15.14\% & 100
& 21.44 & 21.03\% & 38.16 & 18.33\% & 100
& 8.34 & 7.12\% & 12.02 & 12.48\% & 100
& 15.63 & 23.15\% & 28.10 & 25.64\% & 100 \\
\bottomrule
\end{tabular}
}
\end{table*}

\begin{table*}[t]
\centering
\scriptsize
\setlength{\tabcolsep}{1.2pt}
\caption{Detailed held-out-domain comparison with PathWise. Each problem size
reports the mean objective Cost and mean Gap (\%) over five independent runs.
Eval is the average number of design-time evaluator calls per run. Lower Gap is
better.}
\label{tab:pathwise-detailed-held-out}
\resizebox{\textwidth}{!}{%
\begin{tabular}{l*{21}{c}}
\toprule
& \multicolumn{7}{c}{OP-ACO ($\uparrow$)}
& \multicolumn{7}{c}{MKP-ACO ($\uparrow$)}
& \multicolumn{7}{c}{BPP-ACO ($\downarrow$)} \\
\cmidrule(lr){2-8} \cmidrule(lr){9-15} \cmidrule(lr){16-22}
Method
& \multicolumn{2}{c}{N=50} & \multicolumn{2}{c}{N=100} & \multicolumn{2}{c}{N=200} & Eval
& \multicolumn{2}{c}{N=100} & \multicolumn{2}{c}{N=200} & \multicolumn{2}{c}{N=300} & Eval
& \multicolumn{2}{c}{N=500} & \multicolumn{2}{c}{N=1000} & \multicolumn{2}{c}{N=2000} & Eval \\
\cmidrule(lr){2-3} \cmidrule(lr){4-5} \cmidrule(lr){6-7}
\cmidrule(lr){9-10} \cmidrule(lr){11-12} \cmidrule(lr){13-14}
\cmidrule(lr){16-17} \cmidrule(lr){18-19} \cmidrule(lr){20-21}
& Cost & Gap & Cost & Gap & Cost & Gap &
& Cost & Gap & Cost & Gap & Cost & Gap &
& Cost & Gap & Cost & Gap & Cost & Gap & \\
\midrule
\multicolumn{22}{c}{\textbf{LLM-based AHD: DeepSeek-V4-Flash}} \\
\midrule
EOH
& 15.06 & 6.07\% & 30.10 & 9.45\% & 53.82 & 14.02\% & 100
& 17.01 & 0.92\% & 43.32 & 2.61\% & 55.81 & 1.82\% & 100
& 204.49 & 1.91\% & 408.44 & 2.01\% & 817.48 & 1.97\% & 100 \\
MCTS-AHD
& 15.14 & 5.53\% & 30.27 & 8.93\% & 54.30 & 13.25\% & 100
& 16.94 & 1.31\% & 42.65 & 4.13\% & 55.57 & 2.45\% & 100
& 204.32 & 1.83\% & 407.97 & 1.89\% & 816.45 & 1.84\% & 100 \\
ReEvo
& 14.95 & 6.69\% & 29.12 & 12.37\% & 51.16 & 18.26\% & 100
& 16.96 & 1.25\% & 43.14 & 3.01\% & 55.78 & 1.85\% & 100
& 204.93 & 2.13\% & 409.37 & 2.24\% & 819.64 & 2.24\% & 100 \\
\methodname{}
& 14.95 & 6.74\% & 29.21 & 12.10\% & 50.78 & 18.84\% & 19.0
& 17.02 & 0.86\% & 43.30 & 2.66\% & 55.79 & 1.91\% & 15.2
& 206.32 & 2.82\% & 412.53 & 3.03\% & 826.71 & 3.12\% & 26.8 \\
PathWise
& 14.85 & 7.30\% & 28.37 & 14.64\% & 48.25 & 22.89\% & 100
& 16.98 & 1.07\% & 43.37 & 2.49\% & 55.79 & 1.81\% & 100
& 205.45 & 2.39\% & 410.18 & 2.44\% & 820.84 & 2.39\% & 100 \\
\midrule
\multicolumn{22}{c}{\textbf{RL Model: Qwen3-4B-Instruct-2507}} \\
\midrule
CALM
& 14.99 & 6.47\% & 29.81 & 10.29\% & 53.18 & 15.04\% & 150
& 16.91 & 1.52\% & 42.91 & 3.52\% & 55.41 & 2.43\% & 150
& 209.25 & 4.28\% & 419.31 & 4.72\% & 842.50 & 5.09\% & 150 \\
\rowcolor{gray!15}\methodname{}
& 15.17 & 5.34\% & 30.36 & 8.62\% & 54.56 & 12.84\% & 13.6
& 17.03 & 0.84\% & 43.52 & 2.17\% & 56.19 & 1.19\% & 12.4
& 206.05 & 2.69\% & 411.62 & 2.80\% & 824.26 & 2.81\% & 11.7 \\
\rowcolor{gray!15}\methodname{} w/SR
& 15.36 & 4.18\% & 30.97 & 6.60\% & 55.84 & 10.47\% & 100
& 17.06 & 0.67\% & 43.84 & 1.44\% & 56.57 & 0.59\% & 100
& 204.21 & 1.77\% & 407.93 & 1.88\% & 816.94 & 1.90\% & 100 \\
\bottomrule
\end{tabular}
}
\end{table*}

\subsection{Design-Set Evaluations to Target Quality}
\label{appendix:train-side-calls-to-target}

To isolate search efficiency from final held-out validation quality, we measure
how many design-time heuristic evaluations the fixed-workflow AHD baselines
require to reach the final design-set objective achieved by \methodname{}.
Table~\ref{tab:train-side-calls-to-target-summary} shows the main trend: the
short \methodname{} reaches its target in 11.7--13.6 evaluations, whereas the
fixed-workflow baselines generally require substantially more evaluations.
Design-set performance is not identical to held-out validation performance,
which is the final reporting split in our combinatorial experiments. However,
because all methods use the same design instances, objective, and evaluator
protocol, design-set target attainment provides a consistent approximate proxy
for their relative convergence efficiency. Among the 15 reported
baseline--domain comparisons, 14 require at least twice as many evaluations as
\methodname{} to reach the same design-set target; the only exception is
MCTS-AHD on BPP-ACO. This near-universal increase in evaluations provides
direct evidence that \methodname{} converges more efficiently. The main-paper
tables separately report held-out validation performance.

\begin{table*}[t]
\centering
\small
\setlength{\tabcolsep}{5pt}
\caption{\textbf{Design-set evaluations to the AHD Agent target.}
For each domain, the AHD Agent column reports its average number of design-time
heuristic evaluations, which produces the target design-set objective. The
remaining columns report the mean first-hitting FE required by
DeepSeek-V4-Flash fixed-workflow baselines to reach the same target, followed
in parentheses by the ratio to the AHD Agent evaluation count. Lower is
better. All baselines reach the target in all five runs except EOH on MKP-ACO
(four of five); full success rates and standard deviations are reported in
Table~\ref{tab:train-side-calls-to-target-detailed}.}
\label{tab:train-side-calls-to-target-summary}
\resizebox{0.92\textwidth}{!}{%
\begin{tabular}{lcccc}
\toprule
Domain & AHD Agent & EOH & MCTS-AHD & ReEvo \\
\midrule
TSP-ACO
& \cellcolor{gray!15}11.8 ($1.0\times$)
& 65.8 ($5.6\times$)
& 62.0 ($5.3\times$)
& 28.4 ($2.4\times$) \\
CVRP-ACO
& \cellcolor{gray!15}12.9 ($1.0\times$)
& 151.0 ($11.7\times$)
& 127.6 ($9.9\times$)
& 137.2 ($10.6\times$) \\
OP-ACO
& \cellcolor{gray!15}13.6 ($1.0\times$)
& 163.2 ($12.0\times$)
& 99.6 ($7.3\times$)
& 60.8 ($4.5\times$) \\
MKP-ACO
& \cellcolor{gray!15}12.4 ($1.0\times$)
& 207.0 ($16.7\times$; 4/5)
& 175.6 ($14.2\times$)
& 178.0 ($14.4\times$) \\
BPP-ACO
& \cellcolor{gray!15}11.7 ($1.0\times$)
& 121.8 ($10.4\times$)
& 9.8 ($0.8\times$)
& 52.8 ($4.5\times$) \\
\bottomrule
\end{tabular}
}
\end{table*}

Detailed success rates and first-hitting FE statistics are reported in
Table~\ref{tab:train-side-calls-to-target-detailed} for both the short
\methodname{} target and the stronger \methodname{} w/SR target.
For the short target, EOH, MCTS-AHD, and ReEvo succeed in 24/25, 25/25,
and 25/25 runs, respectively, but require 139.0, 94.9, and 91.4 FE on
average over successful runs. For the \methodname{} w/SR target, the
corresponding success counts fall to 2/25, 17/25, and 20/25, and only
three of the 15 baseline--domain configurations reach the target in all
five runs.

\label{appendix:train-side-calls-to-target-detailed}
\begin{table*}[t]
\centering
\scriptsize
\setlength{\tabcolsep}{2.2pt}
\caption{\textbf{Detailed design-set calls to target quality.}
Each baseline cell reports the number of successful runs out of five, followed
by the mean and standard deviation of the first-hitting heuristic evaluation
count over successful runs. Target Eval is the average total design-time
evaluation count of AHD Agent or AHD Agent w/SR. A dash indicates that no run
reaches the target within the extended evaluation horizon. All measurements
use design-set best-so-far trajectories; held-out validation feedback is not
used. Lower FE is better.}
\label{tab:train-side-calls-to-target-detailed}
\resizebox{\textwidth}{!}{%
\begin{tabular}{lcccccccc}
\toprule
& \multicolumn{4}{c}{Target: AHD Agent}
& \multicolumn{4}{c}{Target: AHD Agent w/SR} \\
\cmidrule(lr){2-5} \cmidrule(lr){6-9}
Domain
& Target Eval & EOH & MCTS-AHD & ReEvo
& Target Eval & EOH & MCTS-AHD & ReEvo \\
\midrule
TSP-ACO
& \cellcolor{gray!15}11.8
& 5/5; $65.8\pm16.0$
& 5/5; $62.0\pm68.0$
& 5/5; $28.4\pm27.9$
& \cellcolor{gray!15}100
& 1/5; 47.0
& 5/5; $135.8\pm104.9$
& 5/5; $91.4\pm39.9$ \\
CVRP-ACO
& \cellcolor{gray!15}12.9
& 5/5; $151.0\pm62.9$
& 5/5; $127.6\pm68.7$
& 5/5; $137.2\pm50.3$
& \cellcolor{gray!15}100
& 1/5; 311.0
& 4/5; $243.5\pm76.6$
& 3/5; $182.7\pm66.8$ \\
OP-ACO
& \cellcolor{gray!15}13.6
& 5/5; $163.2\pm102.3$
& 5/5; $99.6\pm17.2$
& 5/5; $60.8\pm37.6$
& \cellcolor{gray!15}100
& 0/5; --
& 3/5; $235.0\pm91.2$
& 4/5; $307.0\pm69.4$ \\
MKP-ACO
& \cellcolor{gray!15}12.4
& 4/5; $207.0\pm71.2$
& 5/5; $175.6\pm89.5$
& 5/5; $178.0\pm80.2$
& \cellcolor{gray!15}100
& 0/5; --
& 5/5; $236.4\pm94.8$
& 4/5; $306.8\pm29.3$ \\
BPP-ACO
& \cellcolor{gray!15}11.7
& 5/5; $121.8\pm109.7$
& 5/5; $9.8\pm5.2$
& 5/5; $52.8\pm79.0$
& \cellcolor{gray!15}100
& 0/5; --
& 0/5; --
& 4/5; $254.0\pm98.1$ \\
\bottomrule
\end{tabular}
}
\end{table*}





\section{External Resources and Licenses}
\label{appendix:resource-licenses}

Table~\ref{tab:appendix-resource-licenses} lists the external codebases, models,
and solver resources referenced or used in our implementation and experiments.
The table is intended as a practical resource summary rather than a complete
legal attribution list.

\begin{table*}[t!]
\centering
\small
\setlength{\tabcolsep}{4pt}
\caption{External resources and license information.}
\label{tab:appendix-resource-licenses}
\resizebox{\textwidth}{!}{
\begin{tabular}{llll}
\toprule
Resource & Type & License & URL \\
\midrule
Unsloth & Code & Apache-2.0 License & \url{https://github.com/unslothai/unsloth} \\
Qwen2.5 & Model & Apache-2.0 License & \url{https://huggingface.co/Qwen/Qwen2.5-7B-Instruct} \\
Optuna & Code & MIT License & \url{https://github.com/optuna/optuna} \\
LKH3 & Code & Available for academic research use & \url{http://webhotel4.ruc.dk/~keld/research/LKH-3/} \\
OR-Tools & Code & MIT License & \url{https://developers.google.com/optimization/pack/knapsack?hl=zh-cn} \\
POMO & Code & Available online & \url{https://github.com/yd-kwon/POMO/tree/master} \\
DeepACO & Code & MIT License & \url{https://github.com/henry-yeh/DeepACO} \\
FunSearch & Code & Apache License & \url{https://github.com/google-deepmind/funsearch} \\
EoH & Code & MIT License & \url{https://github.com/FeiLiu36/EoH/tree/main} \\
ReEvo & Code & MIT License & \url{https://github.com/ai4co/reevo} \\
HSEvo & Code & Available online & \url{https://github.com/datphamvn/HSEvo} \\
MCTS-AHD & Code & MIT License & \url{https://github.com/zz1358m/MCTS-AHD-master} \\
\bottomrule
\end{tabular}
}
\end{table*}

\end{document}